# Numerical Comparison of Neighbourhood Topologies in Particle Swarm Optimization

Mauro S. Innocente[*] and Johann Sienz[†]

*ADOPT Research Group, Civil and Computational Engineering Centre (C²EC), College of Engineering, Swansea University, Singleton Park, Swansea, SA2 8PP, UK*

**Particle Swarm Optimization is a global optimizer in the sense that it has the ability to escape poor local optima. However, if the spread of information within the population is not adequately performed, premature convergence may occur. The convergence speed and hence the reluctance of the algorithm to getting trapped in suboptimal solutions are controlled by the settings of the coefficients in the velocity update equation as well as by the neighbourhood topology. The coefficients settings govern the trajectories of the particles towards the good locations identified, whereas the neighbourhood topology controls the form and speed of spread of information within the population (i.e. the update of the social attractor). Numerous neighbourhood topologies have been proposed and implemented in the literature. This paper offers a numerical comparison of the performances exhibited by five different neighbourhood topologies combined with four different coefficients' settings when optimizing a set of benchmark unconstrained problems. Despite the optimum topology being problem-dependent, it appears that dynamic neighbourhoods with the number of interconnections increasing as the search progresses should be preferred for a non-problem-specific optimizer.**

## Nomenclature

| | | |
|---|---|---|
| C-PSO | = | Constricted Particle Swarm Optimization |
| EAs | = | Evolutionary Algorithms |
| $i$D | = | $i$-dimensional |
| LHS | = | Latin Hypercube Sampling |
| $m$ | = | Swarm size |
| $nn$ / $nni$ / $nnf$ | = | Number of neighbours / initial number of neighbours / final number of neighbours |
| PB_ME | = | Position-Based Mean Error (measure of degree of clustering) |
| PSO | = | Particle Swarm Optimization |
| PSO-RRR$i$ | = | PSO with Reduced Randomness Range $i$ |
| $tref$ | = | Number of time-steps considered in the average of the degree of clustering measures |

## I.Introduction

The Particle Swarm Optimization (PSO) method is considered a global optimizer because it has the ability to escape local optima. However, if the spread of information is not appropriately carried out, premature convergence is possible. The latter does not necessarily take place at a local optimum but at any location, as long as the convergence condition is met (i.e. velocity equal to zero).

The speed of convergence of a given particle to a given attractor is governed by the settings of the coefficients in the velocity update equation. The speed of convergence of the whole algorithm –and hence its reluctance to getting trapped in suboptimal solutions– also depends on the speed of spread of information throughout the swarm. The latter is governed by the neighbourhood topology. In the original algorithm as presented in Ref. 1, the topology was global. This means that every particle is connected to every other, and therefore has access to the best solution found by any particle in the swarm. This leads to the fastest spread of acquired information throughout the swarm, which may result in premature convergence due to premature loss of diversity. To the best of our knowledge, the first local version of the algorithm was proposed in Ref. 2, followed by numerous neighbourhood topologies proposed and implemented in the literature. Given that the connections are commonly defined only once, at the beginning, neighbours are not necessarily near one another in the physical space. Nonetheless, the latter can be done for some topologies at the expense of some additional computational cost. It is evident that the fewer the interconnections the slower the spread of information and therefore the slower the convergence. Since the optimum speed of convergence is problem-dependent, it can be inferred that the optimum neighbourhood topology is also problem-dependent.

---

[*] Post-doc Research Assistant, College of Engineering, M.S.Innocente@swansea.ac.uk.

[†] Professor, Aerospace Research Theme Leader and Portfolio Director, Co-Director of the Welsh Composites Centre (WCC) & Director of ASTUTE project, College of Engineering, J.Sienz@swansea.ac.uk.

Other means to control the spread of information is the number of sociality terms in the velocity update equation. The classical PSO algorithm has only one sociality term, which is either the global best (global topology) or the local best (local topologies). However, it is not difficult to imagine that the particle may be attracted towards the experiences of other particles in the swarm as well. A straightforward variation is to have two sociality terms, one being the best experience in the particles' neighbourhood and the other best in the whole swarm. This can be viewed as the experiences that are directly observed by social beings, plus the experiences not directly observed but '*stored*' in the form of culture. Another more radical variation is the *fully informed PSO* proposed in Ref. 3, where all neighbours comprise a source of influence.

This paper offers a numerical comparison of the performances exhibited by five different neighbourhood topologies combined with four different coefficients settings when optimizing a set of benchmark unconstrained problems. Section II is devoted to the particle swarm optimization paradigm, where the velocities and positions updates and the neighbourhood topologies are discussed in different subsections. Experimental studies are provided in section III, and final conclusions are offered in section IV.

## II. Particle Swarm Optimization

The PSO method was invented by social-psychologist James Kennedy and electrical-engineer Russell C. Eberhart in 1995[1]. It is a population-based, gradient-free method inspired by earlier bird flock simulations and other studies of social behaviour (e.g. Refs. 4 and 5), while also being influenced by previous Evolutionary Algorithms (EAs). The function to be minimized is commonly called '*fitness function*', term imported from the EAs literature. It is more appropriately referred to as '*conflict function*' hereafter due to the social-psychology metaphor that inspired the method.

Gradient information is not required, which enables the method to deal with non-differentiable and even discontinuous problems. Therefore there is no restriction to the characteristics of the *conflict function* for the approach to be applicable. In fact, the function does not even need to be explicit.

Since the method is not designed to optimize but to carry out procedures that are not directly related to the optimization problem, it is frequently referred to as a *modern heuristics*. Optimization occurs without evident links between the implemented procedures and the resulting optimization process: the ability of the swarm to optimize emerges from the cooperation among its particles. While this makes it difficult to understand how optimization takes place, the method shows astonishing robustness in dealing with different kinds of complex problems that it was not specifically designed for. Despite the disadvantage that its theoretical bases are difficult to be understood deterministically, considerable theoretical work has been carried out on simplified versions of the algorithm (e.g. Refs. 6, 7, 8, 9, 10, 11, and 12). For a comprehensive review, refer to Refs. 13, 14, and 15.

### A. Velocity and Position Update

While the emergent optimization capabilities of the PSO algorithm result from local interactions among particles in a swarm, the behaviour of a single particle can be summarized in three sequential processes: *evaluation*; *comparison*; and *imitation*. Thus, the performance of a particle in its current position is evaluated in terms of the conflict function. In order to decide upon its next position, the particle compares its current conflict to those associated with its own and with its neighbours' best experiences. Finally, the particle imitates its own best experience and the best experience in its neighbourhood to some extent. The basic update equations are:

$$\begin{cases} v_{ij}^{(t)} = w \cdot v_{ij}^{(t-1)} + iw \cdot U_{(0,1)} \cdot \left( pbest_{ij}^{(t-1)} - x_{ij}^{(t-1)} \right) + sw \cdot U_{(0,1)} \cdot \left( lbest_{ij}^{(t-1)} - x_{ij}^{(t-1)} \right) \\ x_{ij}^{(t)} = x_{ij}^{(t-1)} + v_{ij}^{(t)} \end{cases} \quad (1)$$

where

$v_{ij}^{(t)}$:     $j^{\text{th}}$ component of the velocity of particle *i* at time-step *t*;

$x_{ij}^{(t)}$:     $j^{\text{th}}$ coordinate of the position of particle *i* at time-step *t*;

$U_{(0,1)}$:     random number from a uniform distribution in the range [0,1] resampled anew every time it is referenced;

$w, iw, sw$: inertia, individuality, and sociality weights, respectively;

$pbest_{ij}^{(t)}$:     $j^{\text{th}}$ coordinate of best position found by particle *i* by time-step *t*;

$lbest_{ij}^{(t)}$:     $j^{\text{th}}$ coordinate of best position found by any particle in the neighbourhood of particle *i* by time-step *t*.

As it can be observed, there are three coefficients that govern the dynamics of the swarm: the inertia ($w$), the individuality ($iw$), and the sociality ($sw$) weights. The settings of these coefficients greatly affect the behaviour of the swarm by controlling the form and speed of convergence of each particle towards its attractors.

## B. Neighbourhood Topology

In the original algorithm as presented in Ref. 1, the topology was *global*. In other words, every particle is connected to every other and therefore has access to the best solution found so far by any particle in the swarm. This leads to very fast convergence –that can be controlled to some extent by the settings of the coefficients– which may result in premature loss of diversity, and consequently in premature convergence.

To the best of our knowledge, the first local version of the algorithm was proposed in Ref. 2. From then on, numerous neighbourhood topologies have been proposed and implemented, some of which are shown in Fig. 1. We propose here to use a ring topology with a linearly increasing number of neighbours, so that the speed of spread of information is slower at the early stages of the search and increases at the search progresses.

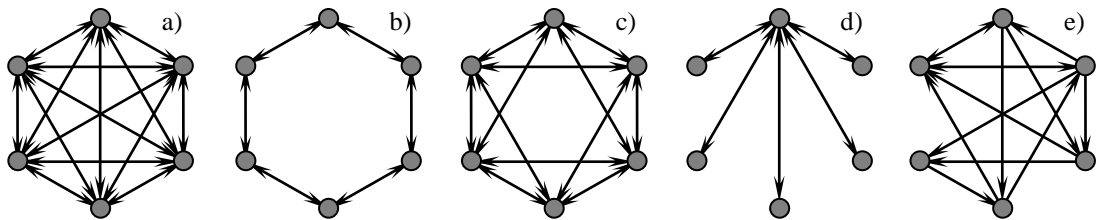

**Figure 1. a) *global* or *fully connected* topology, where neighbourhood size equals swarm size; b) *ring* topology with neighbourhood size equal to three (two neighbours per particle); c) *ring* topology with neighbourhood size equal to five (four neighbours per particle); d) *wheel* topology, where neighbourhood size equals swarm size for one particle and two for the rest; e) *random* topology, where the average neighbourhood size is approximately equal to half the swarm size.**

Clerc[14, p.49] suggests neighbourhoods composed of three particles as a general setting for a standard PSO algorithm with fixed neighbourhoods, while he offers a fairly extensive discussion on the graphs of influence later on in his book[14, pp. 87-101]. Engelbrecht[13, pp. 178-188] also discusses the problem of neighbourhood topologies.

The first dynamic topology in PSO –to the best of our knowledge– was proposed by Suganthan[16], where the social attractor varied from the local to the swarm's best experience. He proposed an empirical formula to calculate a threshold, where the neighbourhood is global if the threshold is passed whereas otherwise the neighbourhood is composed of the particles within a given normalized distance. Another dynamic topology is the '*Stochastic Star*' topology proposed in Ref. 17: at each time-step and for each dimension of the search-space, there is a probability that a particle will not access the global best information, and therefore would move only under the influence of the inertia and the attraction towards its individual best experience. A similar approach to the dynamic neighbourhood proposed here is that in Ref. 18, where each particle starts accessing only one other particle's best experience, and the number of neighbours is increased at regular intervals until it becomes global by the time 80% of the search length has elapsed.

Other classical topologies that have not been tested in this paper are the '*von Neumann*' and the '*Pyramid*' topologies (refer to Refs. 13 pp. 107-109, and 19). For further studies on neighbourhood topologies, refer to Refs. 12, 14 pp. 87-101, 20, 21, 22, 23, 24, 25, and 26.

Given that the connections are commonly defined only once, at the beginning, neighbours are not necessarily near one another in the physical space. As previously mentioned, the fewer the interconnections the slower the convergence, while the problem-dependent optimum speed of convergence makes the optimum neighbourhood topology also problem-dependent. Nevertheless, if a given problem space is to be searched for which there is no information available with regards to its landscape, it makes sense to scout the space first in order to identify potential good areas before spending too much exploiting effort. Hence it seems evident that dynamic neighbourhoods should be preferred if the optimizer is not meant to be problem-specific, increasing the number of interconnections as the search progresses.

## III. Experimental study

Different settings of the coefficients in the velocity update equation notably affect the behaviour of the swarm, as they govern the form and speed of convergence of each particle towards a randomly weighted average of its individual and social attractors. In turn, the neighbourhoods' structure governs the form and speed of spread of individually acquired information throughout the population, thus governing the update of every particle's social attractor. Therefore, the coefficients' settings and the neighbourhood topology together control the speed and form of convergence of the algorithm as a whole. In addition to that, two settings with similar speeds of convergence may result in very different forms of exploration of the search-space: for instance, consider the following two cases, which may lead to similar convergence speeds:

o coefficients settings that favour fast convergence coupled with a neighbourhood structure with few interconnections

o coefficients settings that delay convergence coupled with a neighbourhood structure that favours fast spread of information

In the first case, each particle quickly converges towards its attractors, thus performing a fine-grain search in a small region around them. Convergence is delayed because the information regarding better social attractors takes longer to become available to the particle in question. This means that not only does the particle converge faster to its current attractors, but also that its social attractor is likely to stay stationary for longer thus giving the particle more time to converge towards it.

In the second case, the information about better social attractors is available to the particle sooner, but the particle takes longer to converge towards its current attractors. As a result, the social attractor is updated before the particle had time to carry out a fine-grain search around it. However this also means that a better exploration of the search space is performed. If the coefficients settings are such that a fine-grain search is delayed but eventually performed, this means that the fine-grain search would only be carried out around social attractors that stay stationary for a long period of time (possibly the global optimum).

As explained with these two examples, not only the speed of convergence has an impact in the performance of the algorithm, but also the way the search space is explored. To take this into account, experiments on five different neighbourhood topologies were performed for four different coefficients settings. These settings not only involve different values of the coefficients in *classical PSO*, but also some modifications to the formulation of the update equations in the PSO algorithm. For further details, refer to Ref. 12.

## C. Coefficients settings for the experiments

In order to account for the classical PSO formulation as in Eq. (1); the *Type 1"* constricted PSO proposed by Clerc and Kennedy[10]; and the PSO with Reduced Range of Randomness proposed by Innocente[12]; the reformulation of the PSO update equations reproduced in Eq. (2) was proposed by Innocente[12].

$$
\begin{cases}
v_{ij}^{(t)} = w \cdot v_{ij}^{(t-1)} + \phi_i \cdot \left( pbest_{ij}^{(t-1)} - x_{ij}^{(t-1)} \right) + \phi_s \cdot \left( lbest_{ij}^{(t-1)} - x_{ij}^{(t-1)} \right) \\
\phi_i = ip \cdot \left[ \phi_{\min} + \left( \phi_{\max} - \phi_{\min} \right) \cdot U_{(0,1)} \right] \\
\phi_s = sp \cdot \left[ \phi_{\min} + \left( \phi_{\max} - \phi_{\min} \right) \cdot U_{(0,1)} \right] \\
ip \in [0,1) \quad ; \quad sp = 1 - ip \\
x_{ij}^{(t)} = x_{ij}^{(t-1)} + v_{ij}^{(t)}
\end{cases}
\tag{2}
$$

### 1. Classical PSO

To translate the proposed reformulation into the classical one, replace $\phi_{\min}$ in Eq. (2) by Eq. (3). Other relations between the two formulations are offered in Eq. (4).

$$
\phi_{\min} = 0
\tag{3}
$$

$$
\left. \begin{array}{l} iw = ip \cdot \phi_{\max} \\ sw = sp \cdot \phi_{\max} \end{array} \right\} \quad \Rightarrow \quad \begin{cases} \phi_i = iw \cdot U_{(0,1)} = U_{(0,iw)} \\ \phi_s = sw \cdot U_{(0,1)} = U_{(0,sw)} \end{cases}
\tag{4}
$$

### 2. Constricted PSO (Type 1")

User selects $aw$ and $\kappa \in (0,1)$, preferably $aw > 4$ (slightly) and $\kappa \to 1$.

$$
\chi = \begin{cases} \dfrac{2 \cdot \kappa}{aw - 2 + \sqrt{aw^2 - 4 \cdot aw}} & \text{if } aw \geq 4 \\ \kappa & \text{otherwise} \end{cases}
\tag{5}
$$

$$
\begin{array}{l}
w = \chi \\
\phi_{\max} = \chi \cdot aw \\
\phi_{\min} = 0
\end{array}
\tag{6}
$$

The constricted PSO settings used in the experiments are as follows:
C-PSO-1:     $\kappa = 0.99994$;     $aw = 4.10$;     $ip = 0.50$.

3. *PSO with Reduced Randomness Range 1*

   User selects $aw \in (1.00, 2.00)$, preferably $1.30 \leq aw \leq 1.80$.

$$w = aw - 1$$
$$\phi_{\max} = \frac{3}{2} \cdot (w + 1)$$
$$\phi_{\min} = \frac{1}{2} \cdot (w + 1)$$

(7)

The PSO-RRR1 settings used in the experiments are as follows:
PSO-RRR1-1:     $aw = 1.80$;     $ip = 0.50$.

4. *PSO with Reduced Randomness Range 2*

   User selects $aw \in (1.000, 2.618)$, preferably $1.70 \leq aw \leq 2.40$.

$$w = \frac{1}{aw} - 2 + aw$$
$$\phi_{\max} = 2 \cdot (w + 1)$$
$$\phi_{\min} = 2 \cdot aw - \phi_{\max}$$

(8)

The PSO-RRR2 settings used in the experiments are as follows:
PSO-RRR2-1:     $aw = 2.40$;     $ip = 0.50$.

## D.  Neighbourhood topologies for the experiments

The neighbourhood topologies considered in the experiments are the *global topology*; a *ring topology with two neighbours*; the *wheel topology*; a *random topology* with an average number of neighbours equal to half the swarm size; and a *dynamic ring topology* with increasing number of neighbours. The *dynamic ring topology* aims to combine the robustness associated with the high *degree of locality* of the *ring topology with 2 neighbours* with the speed of convergence and ability to perform fine-grain search of the *global topology*. For the *random topology*, the number of neighbours ($nn$) for each particle at each time-step is randomly generated, and then $nn$ particles from the swarm are randomly chosen as its neighbours.

## E.  Other settings for the experiments

The so-called '*position-based mean error*' (**pb_me**) is the average in a given number of passed time-steps (*tref*) of the square root of the average (among all particles) of the squared normalized (with respect to the feasible intervals and to the number of dimensions) distance between each particle and the best solution:

$$\mathrm{pb\_me}^{(t)} = \frac{\displaystyle\sum_{i=t-tref+1}^{t} \sqrt{\dfrac{\displaystyle\sum_{j=1}^{n}\sum_{k=1}^{m}\left(\dfrac{x_{kj}^{(i)} - gbest_{j}^{(i)}}{x_{j\max} - x_{j\min}}\right)^{2}}{m \cdot n}}}{tref} = \frac{\displaystyle\sum_{i=t-tref+1}^{t} \sqrt{\displaystyle\sum_{j=1}^{n}\dfrac{\displaystyle\sum_{k=1}^{m}\left(x_{kj}^{(i)} - gbest_{j}^{(i)}\right)^{2}}{m \cdot \left(x_{j\max} - x_{j\min}\right)^{2}}}}{tref \cdot \sqrt{n}}$$

(9)

The PSO-RRR2-1, PSO-RRR1-1, and the C-PSO-1 previously described, plus a Multi-Swarm algorithm combining the three of them, are tested on a well-known suite of unconstrained benchmark problems composed of the *Sphere, Rosenbrock, Rastrigin, Griewank,* and *Schaffer f6* problems, each one in 2D, 10D, and 30D. Refer to Ref. 12 for their formulations. Every run is performed with a swarm of 50 particles for a length of 10000 time-steps. Intermediate results at 1000[th] time-step are also provided.

The particles' positions (**p**) are initialized by generating 1000 independent Latin Hypercube Samplings (LHSs), and selecting the one with the maximum minimum distance between particles. Velocities are initialized to zero, and the individual best experiences (**pbest**) are initialized instead. Every best experience is initialized at exactly the same distance from its corresponding particle. Each component of this distance is calculated as the corresponding feasible interval divided by twice the number of particles in the swarm. The sign of the component, and hence the direction of the distance vector, are randomly generated. For each pair '**p–pbest**', a comparison is performed so that the best one becomes (or stays) **pbest** and the other becomes (or stays) **p** before the search begins. Thus, every particle starts the search with the same, moderate acceleration towards its **pbest** (the acceleration towards its **lbest** will depend on the neighbourhood structure). Therefore, in the end, the particles' initialization will most likely not be a LHS. For the Multi-Swarm algorithm, each sub-swarm is initialized independently.

Interval constraints are simply treated as any other inequality constraint, and infeasible particles are not evaluated. This way, the normal dynamics of the swarm is least disrupted, and the particles could approach the solution from every direction without losing momentum too quickly when the solution lies near the boundaries.

A run is considered successful if the error is no greater than $10^{-4}$, and the statistics are calculated out of 25 runs. The random number generator is reset to its initial state only before the first run of every experiment.

## F. Experimental results

The results obtained from the experiments are provided in Table 1 to Table 15, and in Fig. 2 to Fig. 28.

### 1. Sphere

In the 2D problem, every combination of coefficients settings and neighbourhood topology finds the exact solution for every run. It can also be observed that the implosion of the particles is virtually complete in every case by the end of the search, while the PSO-RRR1-1 approach exhibits the highest degree of clustering by the $1,000^{th}$ time-step regardless of the neighbourhood topology (see pb_me in Table 1).

In the 10D problem, every algorithm achieves a 100% success rate (SR) by the end of the search. It can also be observed that the 'PSO-RRR2-1 Random' is the only one whose mean solution does not meet the success criterion (i.e. error below $10^{-4}$) by the $1,000^{th}$ time-step. It is also the one with the lowest degree of clustering by the end of the search. The values of pb_me in Table 2 show that the PSO-RRR1-1 reaches the highest degree of clustering, the PSO-RRR2-1 the lowest, and the C-PSO-1 and the Multi-Swarm (MS) are in between.

In the 30D problem, the 'PSO-RRR2-1 Random' and the 'PSO-RRR1-1 Global' are the only ones which do not achieve a SR of 100% by the end of the search, but for two very different reasons: the former because convergence is too slow, and the latter because of premature convergence (refer to Fig. 4). The dynamic neighbourhood appears successful: the median and mean solutions found are either between those obtained by the global and the ring ($nn$=2) topologies (PSO-RRR2-1, C-PSO-1), or they are better than both (PSO-RRR1-1, MS). Making the neighbourhood dynamic avoids the premature convergence observed in the 'PSO-RRR1-1 Global', while achieving better solutions and higher clustering than the 'PSO-RRR1-1 Ring $nn$=2' (see Table 3).

### 2. Rosenbrock

In the 2D problem, every algorithm finds the exact solution for every run. It can also be observed that the implosion of the particles is virtually complete in every case by the end of the search, while the PSO-RRR1-1 approach presents the highest degree of clustering by the $1,000^{th}$ time-step (see pb_me in Table 4).

In the 10D problem, the success rates (SRs) of most algorithms fall dramatically. In the same fashion as when optimizing the 30D Sphere, the 'PSO-RRR2-1 Random' and the 'PSO-RRR1-1 Global' obtained the worst results, the former due to slow convergence whereas the latter due to premature convergence (Fig. 7). Although the 'PSO-RRR1-1 Global' achieved a SR of 44%, it converged to a local optimum (3.99) 36% of the times. The best performance is exhibited by the 'PSO-RRR1-1 Ring $nn$=2', which is the only algorithm that achieved 100% success. As to the proposed dynamic topology, results are very promising. For the PSO-RRR2-1, the SR and the median solution are between those of the global and of the ring ($nn$=2) topologies (closer to the better one), as expected, while the mean solution is better than both, never falling into the local optimum. For the PSO-RRR1-1, the mean solution and the SR are between those of the global and the ring ($nn$=2) topologies, while the median solution is better than both. The SR is 96%, falling into a local optimum only once in 25 runs. For the C-PSO-1, the median, the mean, and the SR are remarkably better than those obtained by the global and by the ring ($nn$=2) topologies, achieving a SR of 96% and never falling into a local optimum (both the global and ring ($nn$=2) do). The same is true for the MS. As shown in Fig. 7 and Fig. 8, the best mean solutions are found by the 'Ring Dynamic' and the 'Ring $nn$=2' topologies (brown and blue curves).

In the 30D problem, achieving success becomes notably harder. The same as before, the 'PSO-RRR2-1 Random' exhibits extremely slow convergence (never achieved) while the 'PSO-RRR1-1 Global' shows premature convergence (see Fig. 9). It is interesting to observe that, after 10,000 time-steps, only four algorithms reach convergence (see Fig. 9 to Fig. 11): the PSO-RRR1-1 Global', which shows premature convergence, the 'PSO-RRR1-1 Ring Dynamic', the 'PSO-RRR1-1 Random', and the 'C-PSO-1 Global'. The latter three obtain the best results. The 'PSO-RRR2-1 Ring Dynamic' shows marginally worse results than its global and ring ($nn$=2) counterparts (none of which converges). The 'PSO-RRR1-1 Ring Dynamic' converges and shows remarkably better performance than its global and ring ($nn$=2) counterparts. The 'C-PSO-1 Ring Dynamic' obtains a median solution worse than, and a mean solution in between, those of its global and ring (*nn*=2) counterparts. Note that this algorithm is still far from converging (brown solid line in Fig. 10 and Fig. 11). Finally, the 'MS Ring Dynamic' finds results in between those of its global and ring (*nn*=2) counterparts.

*3. Rastrigin*

In the 2D problem, every algorithm finds the exact solution for every run by the end of the search. In fact, they all do by the 1000[th] time-step already, except for the 'MS Ring *nn*=2' (which also shows a remarkable lower degree of clustering). It is not clear why convergence is so delayed in this case (see Table 7).

In the 10D problem, the 'Ring Dynamic' topologies result in a remarkable increase in the success rate (SR) when compared to their global and ring (*nn*=2) counterparts for every coefficients settings (see Table 8). Since this is a highly multimodal function, several cases of (early) stagnation can be observed in Fig. 13 and Fig. 14. The best performance overall is exhibited by the 'PSO-RRR2-1 Ring Dynamic', while the 'C-PSO-1 Ring Dynamic' and the 'MS Ring Dynamic' also show very good performance.

In the 30D problem, no algorithm is able to meet the success criterion in any run. The best performance is exhibited by the 'C-PSO-1 Random' (see Table 9) although convergence is very slow (see Fig. 15 to Fig. 17). The 'PSO-RRR2-1 Wheel' and 'MS Random' also find very good solutions, while exhibiting faster convergence and a much earlier stagnation. Notice that all the global topologies, the 'PSO-RRR1-1 Wheel', the 'PSO-RRR1-1 Random', and the 'MS Random', show a complete loss of diversity (see the values of pb_me in Table 9) and stagnation (see convergence curves in Fig. 15 to Fig. 17). All the 'Ring *nn*=2' and the 'Ring Dynamic' topologies still present some diversity by the end of the search, so that improvement is to be expected for an extended search-length. Nonetheless, the 'PSO-RRR2-1 Ring Dynamic' and the 'MS Ring Dynamic' exhibit some of the best performances. For the 'PSO-RRR1-1' and the 'C-PSO-1', the 'Ring Dynamic' topology finds solutions in between their global and ring (*nn*=2) counterparts, while the 'MS Ring Dynamic' exhibits better performance than both. Instead, the 'PSO-RRR2-1 Ring Dynamic' shows similar performance to that of the 'PSO-RRR2-1 Ring *nn*=2', which, surprisingly, are worse than that of the global topology in a highly multimodal problem. This is because the combination of PSO-RRR2-1 (slow convergence) with highly local neighbours for a high-dimensional and highly multimodal problem results in too slow a convergence for a search this long. Notice that the ring (*nn*=2) and dynamic topologies did not fully converge, displaying notably higher diversity.

*4. Griewank*

In the 2D problem, three algorithms exhibit premature convergence: the 'PSO-RRR2-1 Global' (in 1 out of 25 runs); the 'PSO-RRR1-1 Global' (in 2 out of 25 runs); and the 'PSO-RRR1-1 Wheel' (in 3 out of 25 runs). In the two global cases, making them dynamic resolves the problem, while at the end of the search they end up showing similar degrees of clustering to those of their global counterparts, and remarkably higher degrees of clustering than those of their ring (*nn*=2) counterparts (see pb_me in Table 10). The other algorithms achieve a 100% success rate (SR).

In the 10D problem, the SRs decrease dramatically. By a large margin, the best performances are exhibited by all the 'Ring *nn*=2' and the 'Ring Dynamic' topologies, as can be clearly seen in Table 11, and in Fig. 19 to Fig. 21. For the PSO-RRR1-1, the 'Ring Dynamic' topology obtains better results than both its global and ring (*nn*=2) counterparts, while in the other cases results are in between, as it would be expected. The best performance overall is exhibited by the 'MS Ring *nn*=2', followed by the 'C-PSO-1 Ring *nn*=2', and the 'C-PSO-1 Ring Dynamic'.

In the 30D problem, the SRs increase, as the difficulty of this particular problem decreases for high dimensionality. By a large margin again, the best performances are exhibited by all the 'Ring *nn*=2' and by the 'Ring Dynamic' topologies, as can be clearly seen in Table 12, and in Fig. 22 and Fig. 23. The performances of the 'Ring Dynamic' neighbourhoods fall between those of their global and ring (*nn*=2) counterparts, as expected, and closer to the better one (i.e. the ring topology). The best performance overall is exhibited by the 'C-PSO-1 Ring *nn*=2', followed by the 'PSO-RRR2-1 Ring *nn*=2'.

*5. Schaffer f6*

In the 2D problem, 9 algorithms, namely all the global and wheel topologies plus the 'C-PSO-1 Ring *nn*=2', present some few failures to achieve the success criterion. The remaining 11 algorithms find the solution in every run (see Table 13 and Fig. 24).

In the 10D problem, no algorithm is able to find the solution in any run (see Table 14). The best performances are exhibited by all the 'Ring Dynamic' neighbourhoods, the 'PSO-RRR2-1 Ring *nn*=2', the 'PSO-RRR1-1 Random', the 'C-PSO-1 Random', and the 'MS Random' (refer to Table 14, and Fig. 25 and Fig. 26).

In the 30D problem, no algorithm is able to find the solution in any run (see Table 15). The best performances are exhibited by the 'PSO-RRR2-1 Ring Dynamic' and the 'C-PSO-1 Ring Dynamic', followed by the 'MS Random' and the 'MS Ring Dynamic' (refer to Table 15 and to Fig. 27 and Fig. 28).

**Table 1. Statistical results out of 25 runs for the PSO-RRR2-1, the PSO-RRR1-1, the C-PSO-1, and a Multi-Swarm algorithm optimizing the 2-dimensional Sphere function. The neighbourhoods tested are the GLOBAL; the RING with 2 neighbours; the RING with linearly increasing number of neighbours (from 2 to 'swarm-size – 1'); the WHEEL; and a RANDOM topology. A run with an error no greater than 0.0001 is regarded as successful.**

| OPTIMIZER | NEIGHBOURHOOD STRUCTURE | | Time-steps | SPHERE 2D | | | | OPTIMUM = 0 | |
|---|---|---|---|---|---|---|---|---|---|
| | | | | BEST | MEDIAN | MEAN | WORST | MEAN PB_ME | [%] Success |
| PSO-RRR2-1 | GLOBAL | | 10000 | 0.00E+00 | 0.00E+00 | 0.00E+00 | 0.00E+00 | 0.00E+00 | 100 |
| | | | 1000 | 1.74E-57 | 3.42E-54 | 4.65E-53 | 5.96E-52 | 4.79E-18 | - |
| | RING | nn = 2 | 10000 | 0.00E+00 | 0.00E+00 | 0.00E+00 | 0.00E+00 | 0.00E+00 | 100 |
| | | | 1000 | 2.61E-53 | 8.64E-49 | 7.07E-46 | 1.26E-44 | 1.55E-18 | - |
| | | nni = 2 nnf = (m – 1) | 10000 | 0.00E+00 | 0.00E+00 | 0.00E+00 | 0.00E+00 | 0.00E+00 | 100 |
| | | | 1000 | 1.76E-54 | 1.58E-50 | 1.80E-48 | 2.05E-47 | 5.87E-19 | - |
| | WHEEL | | 10000 | 0.00E+00 | 0.00E+00 | 0.00E+00 | 0.00E+00 | 0.00E+00 | 100 |
| | | | 1000 | 2.96E-54 | 7.33E-48 | 1.69E-39 | 4.23E-38 | 5.71E-19 | - |
| | RANDOM | | 10000 | 0.00E+00 | 0.00E+00 | 0.00E+00 | 0.00E+00 | 0.00E+00 | 100 |
| | | | 1000 | 1.74E-51 | 3.89E-49 | 2.60E-48 | 1.45E-47 | 2.63E-19 | - |
| PSO-RRR1-1 | GLOBAL | | 10000 | 0.00E+00 | 0.00E+00 | 0.00E+00 | 0.00E+00 | 0.00E+00 | 100 |
| | | | 1000 | 5.19E-88 | 2.30E-85 | 3.33E-84 | 4.17E-83 | 3.43E-37 | - |
| | RING | nn = 2 | 10000 | 0.00E+00 | 0.00E+00 | 0.00E+00 | 0.00E+00 | 0.00E+00 | 100 |
| | | | 1000 | 1.09E-82 | 2.99E-80 | 2.13E-78 | 2.27E-77 | 1.52E-37 | - |
| | | nni = 2 nnf = (m – 1) | 10000 | 0.00E+00 | 0.00E+00 | 0.00E+00 | 0.00E+00 | 0.00E+00 | 100 |
| | | | 1000 | 1.65E-84 | 8.87E-82 | 6.58E-81 | 8.62E-80 | 3.99E-38 | - |
| | WHEEL | | 10000 | 0.00E+00 | 0.00E+00 | 0.00E+00 | 0.00E+00 | 0.00E+00 | 100 |
| | | | 1000 | 4.81E-86 | 3.57E-80 | 1.30E-76 | 3.14E-75 | 1.59E-37 | - |
| | RANDOM | | 10000 | 0.00E+00 | 0.00E+00 | 0.00E+00 | 0.00E+00 | 0.00E+00 | 100 |
| | | | 1000 | 2.01E-86 | 4.41E-83 | 1.33E-82 | 7.29E-82 | 6.36E-37 | - |
| C-PSO-1 | GLOBAL | | 10000 | 0.00E+00 | 0.00E+00 | 0.00E+00 | 0.00E+00 | 0.00E+00 | 100 |
| | | | 1000 | 2.74E-91 | 5.15E-88 | 2.06E-84 | 5.14E-83 | 9.16E-30 | - |
| | RING | nn = 2 | 10000 | 0.00E+00 | 0.00E+00 | 0.00E+00 | 0.00E+00 | 0.00E+00 | 100 |
| | | | 1000 | 5.58E-82 | 3.44E-78 | 5.67E-76 | 6.45E-75 | 3.58E-32 | - |
| | | nni = 2 nnf = (m – 1) | 10000 | 0.00E+00 | 0.00E+00 | 0.00E+00 | 0.00E+00 | 0.00E+00 | 100 |
| | | | 1000 | 7.89E-85 | 1.89E-81 | 2.99E-79 | 4.49E-78 | 4.59E-33 | - |
| | WHEEL | | 10000 | 0.00E+00 | 0.00E+00 | 0.00E+00 | 0.00E+00 | 0.00E+00 | 100 |
| | | | 1000 | 2.41E-87 | 2.54E-79 | 1.85E-67 | 4.63E-66 | 3.51E-33 | - |
| | RANDOM | | 10000 | 0.00E+00 | 0.00E+00 | 0.00E+00 | 0.00E+00 | 0.00E+00 | 100 |
| | | | 1000 | 1.41E-87 | 2.03E-84 | 1.54E-82 | 2.99E-81 | 3.73E-34 | - |
| Multi-Swarm | GLOBAL | | 10000 | 0.00E+00 | 0.00E+00 | 0.00E+00 | 0.00E+00 | 0.00E+00 | 100 |
| | | | 1000 | 2.38E-90 | 4.67E-86 | 1.06E-83 | 2.16E-82 | 1.70E-20 | - |
| | RING | nn = 2 | 10000 | 0.00E+00 | 0.00E+00 | 0.00E+00 | 0.00E+00 | 0.00E+00 | 100 |
| | | | 1000 | 3.39E-85 | 4.26E-78 | 9.85E-77 | 1.83E-75 | 2.68E-20 | - |
| | | nni = 2 nnf = (m – 1) | 10000 | 0.00E+00 | 0.00E+00 | 0.00E+00 | 0.00E+00 | 0.00E+00 | 100 |
| | | | 1000 | 2.99E-86 | 1.12E-80 | 2.13E-79 | 2.33E-78 | 4.47E-20 | - |
| | WHEEL | | 10000 | 0.00E+00 | 0.00E+00 | 0.00E+00 | 0.00E+00 | 0.00E+00 | 100 |
| | | | 1000 | 2.69E-64 | 2.13E-48 | 8.29E-44 | 2.07E-42 | 1.92E-19 | - |
| | RANDOM | | 10000 | 0.00E+00 | 0.00E+00 | 0.00E+00 | 0.00E+00 | 0.00E+00 | 100 |
| | | | 1000 | 1.59E-85 | 2.47E-82 | 2.84E-80 | 5.68E-79 | 2.37E-20 | - |

**Table 2. Statistical results out of 25 runs for the PSO-RRR2-1, the PSO-RRR1-1, the C-PSO-1, and a Multi-Swarm algorithm optimizing the 10-dimensional Sphere function. The neighbourhoods tested are the GLOBAL; the RING with 2 neighbours; the RING with linearly increasing number of neighbours (from 2 to 'swarm-size – 1'); the WHEEL; and a RANDOM topology. A run with an error no greater than 0.0001 is regarded as successful.**

| OPTIMIZER | NEIGHBOURHOOD STRUCTURE | | Time-steps | SPHERE 10D | | | | OPTIMUM = 0 | [%] Success |
|---|---|---|---|---|---|---|---|---|---|
| | | | | BEST | MEDIAN | MEAN | WORST | MEAN PB_ME | |
| PSO-RRR2-1 | GLOBAL | | 10000 | 2.04E-256 | 3.90E-250 | 3.93E-247 | 8.75E-246 | 2.28E-126 | 100 |
| | | | 1000 | 4.97E-24 | 5.49E-23 | 1.25E-22 | 1.06E-21 | 8.08E-14 | - |
| | RING | nn = 2 | 10000 | 1.29E-145 | 2.82E-143 | 3.37E-141 | 5.90E-140 | 1.13E-73 | 100 |
| | | | 1000 | 1.78E-13 | 3.67E-12 | 5.33E-12 | 2.02E-11 | 8.77E-09 | - |
| | | nni = 2 nnf = (m – 1) | 10000 | 1.46E-226 | 5.19E-223 | 5.65E-221 | 6.34E-220 | 1.32E-113 | 100 |
| | | | 1000 | 2.26E-16 | 7.12E-15 | 1.30E-14 | 6.38E-14 | 2.46E-10 | - |
| | WHEEL | | 10000 | 9.84E-173 | 5.18E-161 | 2.52E-154 | 5.85E-153 | 4.05E-81 | 100 |
| | | | 1000 | 1.44E-15 | 2.61E-14 | 1.09E-12 | 1.14E-11 | 6.59E-10 | - |
| | RANDOM | | 10000 | 1.31E-72 | 2.36E-67 | 6.47E-65 | 1.05E-63 | 6.16E-36 | 100 |
| | | | 1000 | 2.31E-06 | 6.63E-05 | 3.40E-04 | 4.98E-03 | 1.98E-05 | - |
| PSO-RRR1-1 | GLOBAL | | 10000 | 0.00E+00 | 0.00E+00 | 0.00E+00 | 0.00E+00 | 0.00E+00 | 100 |
| | | | 1000 | 4.75E-67 | 1.61E-65 | 5.10E-65 | 3.50E-64 | 5.98E-35 | - |
| | RING | nn = 2 | 10000 | 0.00E+00 | 0.00E+00 | 0.00E+00 | 0.00E+00 | 0.00E+00 | 100 |
| | | | 1000 | 4.86E-35 | 2.94E-33 | 4.46E-33 | 1.53E-32 | 1.01E-19 | - |
| | | nni = 2 nnf = (m – 1) | 10000 | 0.00E+00 | 0.00E+00 | 0.00E+00 | 0.00E+00 | 0.00E+00 | 100 |
| | | | 1000 | 4.00E-44 | 7.30E-43 | 2.52E-42 | 1.96E-41 | 1.93E-24 | - |
| | WHEEL | | 10000 | 0.00E+00 | 0.00E+00 | 0.00E+00 | 0.00E+00 | 0.00E+00 | 100 |
| | | | 1000 | 3.62E-48 | 1.39E-43 | 1.02E-41 | 1.74E-40 | 9.63E-25 | - |
| | RANDOM | | 10000 | 0.00E+00 | 0.00E+00 | 0.00E+00 | 0.00E+00 | 0.00E+00 | 100 |
| | | | 1000 | 3.74E-54 | 8.67E-53 | 2.50E-52 | 1.48E-51 | 2.26E-29 | - |
| C-PSO-1 | GLOBAL | | 10000 | 0.00E+00 | 0.00E+00 | 0.00E+00 | 0.00E+00 | 0.00E+00 | 100 |
| | | | 1000 | 1.30E-51 | 6.84E-50 | 3.49E-49 | 5.15E-48 | 7.98E-27 | - |
| | RING | nn = 2 | 10000 | 1.13E-280 | 2.23E-277 | 4.17E-274 | 5.97E-273 | 2.11E-140 | 100 |
| | | | 1000 | 7.36E-27 | 3.12E-25 | 4.91E-25 | 2.77E-24 | 1.50E-15 | - |
| | | nni = 2 nnf = (m – 1) | 10000 | 0.00E+00 | 0.00E+00 | 0.00E+00 | 0.00E+00 | 0.00E+00 | 100 |
| | | | 1000 | 7.32E-33 | 1.59E-31 | 3.09E-31 | 2.69E-30 | 1.18E-18 | - |
| | WHEEL | | 10000 | 0.00E+00 | 0.00E+00 | 0.00E+00 | 0.00E+00 | 0.00E+00 | 100 |
| | | | 1000 | 6.57E-35 | 2.83E-30 | 2.16E-29 | 3.11E-28 | 3.41E-18 | - |
| | RANDOM | | 10000 | 0.00E+00 | 0.00E+00 | 0.00E+00 | 0.00E+00 | 0.00E+00 | 100 |
| | | | 1000 | 1.27E-36 | 3.86E-34 | 9.19E-33 | 1.87E-31 | 1.24E-19 | - |
| Multi-Swarm | GLOBAL | | 10000 | 0.00E+00 | 0.00E+00 | 0.00E+00 | 0.00E+00 | 0.00E+00 | 100 |
| | | | 1000 | 3.21E-58 | 2.00E-56 | 2.03E-55 | 1.42E-54 | 2.17E-18 | - |
| | RING | nn = 2 | 10000 | 0.00E+00 | 0.00E+00 | 0.00E+00 | 0.00E+00 | 1.20E-153 | 100 |
| | | | 1000 | 1.41E-32 | 1.17E-30 | 1.06E-29 | 9.91E-29 | 2.68E-11 | - |
| | | nni = 2 nnf = (m – 1) | 10000 | 0.00E+00 | 0.00E+00 | 0.00E+00 | 0.00E+00 | 0.00E+00 | 100 |
| | | | 1000 | 4.18E-43 | 2.00E-39 | 5.49E-38 | 1.09E-36 | 1.35E-14 | - |
| | WHEEL | | 10000 | 2.19E-202 | 5.77E-193 | 1.51E-186 | 2.67E-185 | 5.31E-97 | 100 |
| | | | 1000 | 6.55E-22 | 4.21E-17 | 2.35E-15 | 3.32E-14 | 2.55E-11 | - |
| | RANDOM | | 10000 | 0.00E+00 | 0.00E+00 | 0.00E+00 | 0.00E+00 | 0.00E+00 | 100 |
| | | | 1000 | 3.71E-44 | 2.13E-42 | 4.52E-41 | 4.89E-40 | 4.41E-17 | - |

**Table 3. Statistical results out of 25 runs for the PSO-RRR2-1, the PSO-RRR1-1, the C-PSO-1, and a Multi-Swarm algorithm optimizing the 30-dimensional Sphere function. The neighbourhoods tested are the GLOBAL; the RING with 2 neighbours; the RING with linearly increasing number of neighbours (from 2 to 'swarm-size – 1'); the WHEEL; and a RANDOM topology. A run with an error no greater than 0.0001 is regarded as successful.**

| OPTIMIZER | NEIGHBOURHOOD STRUCTURE | | Time-steps | SPHERE 30D | | | | OPTIMUM = 0 | |
|---|---|---|---|---|---|---|---|---|---|
| | | | | BEST | MEDIAN | MEAN | WORST | MEAN PB_ME | [%] Success |
| PSO-RRR2-1 | GLOBAL | | 10000 | 1.22E-87 | 3.29E-84 | 3.07E-82 | 6.31E-81 | 2.12E-45 | 100 |
| | | | 1000 | 3.49E-06 | 1.85E-05 | 4.08E-05 | 2.77E-04 | 1.84E-06 | - |
| | RING | nn = 2 | 10000 | 3.77E-43 | 1.90E-42 | 7.86E-42 | 6.78E-41 | 9.24E-25 | 100 |
| | | | 1000 | 1.23E-01 | 2.84E-01 | 3.12E-01 | 6.69E-01 | 2.14E-04 | - |
| | | nni = 2 nnf = (m – 1) | 10000 | 2.21E-74 | 8.42E-73 | 6.49E-72 | 6.44E-71 | 3.87E-40 | 100 |
| | | | 1000 | 9.91E-03 | 2.85E-02 | 2.98E-02 | 7.82E-02 | 4.65E-05 | - |
| | WHEEL | | 10000 | 9.29E-50 | 8.45E-47 | 7.72E-45 | 1.17E-43 | 4.78E-27 | 100 |
| | | | 1000 | 3.04E-02 | 9.25E-02 | 1.34E-01 | 3.69E-01 | 5.88E-05 | - |
| | RANDOM | | 10000 | 6.28E-07 | 5.97E-04 | 2.90E-03 | 2.00E-02 | 1.56E-05 | 20 |
| | | | 1000 | 2.75E+02 | 1.02E+03 | 1.25E+03 | 3.38E+03 | 1.01E-02 | - |
| PSO-RRR1-1 | GLOBAL | | 10000 | 4.06E-07 | 3.79E-04 | 9.89E-02 | 2.39E+00 | 4.69E-17 | 32 |
| | | | 1000 | 5.57E-06 | 1.64E-03 | 2.70E-01 | 4.03E+00 | 5.93E-11 | - |
| | RING | nn = 2 | 10000 | 8.13E-144 | 7.26E-142 | 5.86E-141 | 6.69E-140 | 1.20E-74 | 100 |
| | | | 1000 | 1.73E-11 | 7.83E-11 | 8.50E-11 | 2.40E-10 | 2.05E-09 | - |
| | | nni = 2 nnf = (m – 1) | 10000 | 2.04E-268 | 4.97E-257 | 6.23E-249 | 8.26E-248 | 1.16E-134 | 100 |
| | | | 1000 | 5.25E-17 | 2.48E-16 | 3.07E-16 | 1.05E-15 | 3.12E-12 | - |
| | WHEEL | | 10000 | 4.05E-48 | 3.99E-40 | 3.64E-29 | 9.08E-28 | 1.20E-22 | 100 |
| | | | 1000 | 2.86E-05 | 1.73E-03 | 1.91E-02 | 2.02E-01 | 2.62E-06 | - |
| | RANDOM | | 10000 | 1.19E-286 | 7.00E-282 | 1.74E-276 | 4.28E-275 | 7.38E-143 | 100 |
| | | | 1000 | 7.92E-27 | 7.54E-25 | 1.36E-23 | 2.88E-22 | 2.67E-16 | - |
| C-PSO-1 | GLOBAL | | 10000 | 3.05E-220 | 2.21E-212 | 1.42E-207 | 3.49E-206 | 1.57E-108 | 100 |
| | | | 1000 | 8.53E-20 | 1.09E-17 | 1.04E-16 | 9.10E-16 | 1.33E-12 | - |
| | RING | nn = 2 | 10000 | 5.68E-96 | 1.67E-94 | 1.65E-93 | 3.57E-92 | 7.60E-51 | 100 |
| | | | 1000 | 7.53E-07 | 3.07E-06 | 3.62E-06 | 1.35E-05 | 5.46E-07 | - |
| | | nni = 2 nnf = (m – 1) | 10000 | 3.06E-182 | 6.62E-179 | 1.98E-177 | 3.84E-176 | 3.81E-93 | 100 |
| | | | 1000 | 7.02E-10 | 4.53E-09 | 4.77E-09 | 1.48E-08 | 1.77E-08 | - |
| | WHEEL | | 10000 | 4.49E-100 | 1.31E-95 | 8.45E-91 | 2.09E-89 | 2.29E-50 | 100 |
| | | | 1000 | 1.74E-08 | 9.80E-07 | 4.16E-06 | 2.23E-05 | 1.49E-07 | - |
| | RANDOM | | 10000 | 4.50E-109 | 1.09E-103 | 7.90E-102 | 5.27E-101 | 4.94E-55 | 100 |
| | | | 1000 | 1.49E-08 | 1.67E-07 | 3.64E-07 | 3.06E-06 | 1.30E-07 | - |
| Multi-Swarm | GLOBAL | | 10000 | 1.10E-181 | 4.68E-172 | 1.95E-166 | 4.21E-165 | 7.90E-88 | 100 |
| | | | 1000 | 4.53E-17 | 1.52E-14 | 4.02E-11 | 1.00E-09 | 8.53E-10 | - |
| | RING | nn = 2 | 10000 | 3.13E-113 | 7.02E-109 | 2.72E-107 | 5.96E-106 | 6.93E-57 | 100 |
| | | | 1000 | 2.83E-08 | 9.73E-08 | 2.55E-07 | 3.82E-06 | 1.75E-06 | - |
| | | nni = 2 nnf = (m – 1) | 10000 | 1.28E-185 | 6.46E-180 | 6.80E-173 | 1.68E-171 | 2.58E-91 | 100 |
| | | | 1000 | 7.07E-12 | 1.43E-10 | 3.51E-10 | 3.97E-09 | 1.57E-08 | - |
| | WHEEL | | 10000 | 4.31E-56 | 1.21E-53 | 6.22E-52 | 6.12E-51 | 1.42E-30 | 100 |
| | | | 1000 | 8.25E-04 | 1.66E-02 | 3.24E-02 | 2.36E-01 | 2.49E-05 | - |
| | RANDOM | | 10000 | 9.07E-180 | 1.28E-175 | 2.08E-172 | 3.32E-171 | 1.66E-90 | 100 |
| | | | 1000 | 1.80E-16 | 1.41E-14 | 1.28E-13 | 2.27E-12 | 1.40E-10 | - |

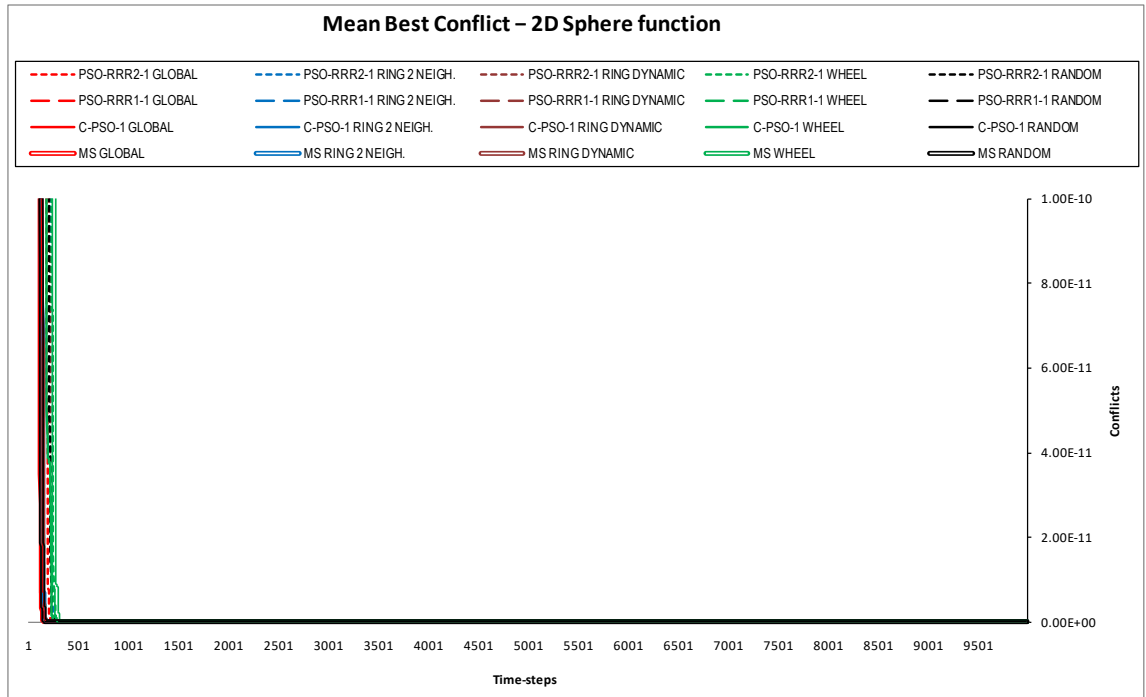

**Figure 2. Convergence curves of the mean best conflict for the 2D Sphere function, associated to Table 1. The colour-codes used to identify the neighbourhood structures are the same in the table and figure associated.**

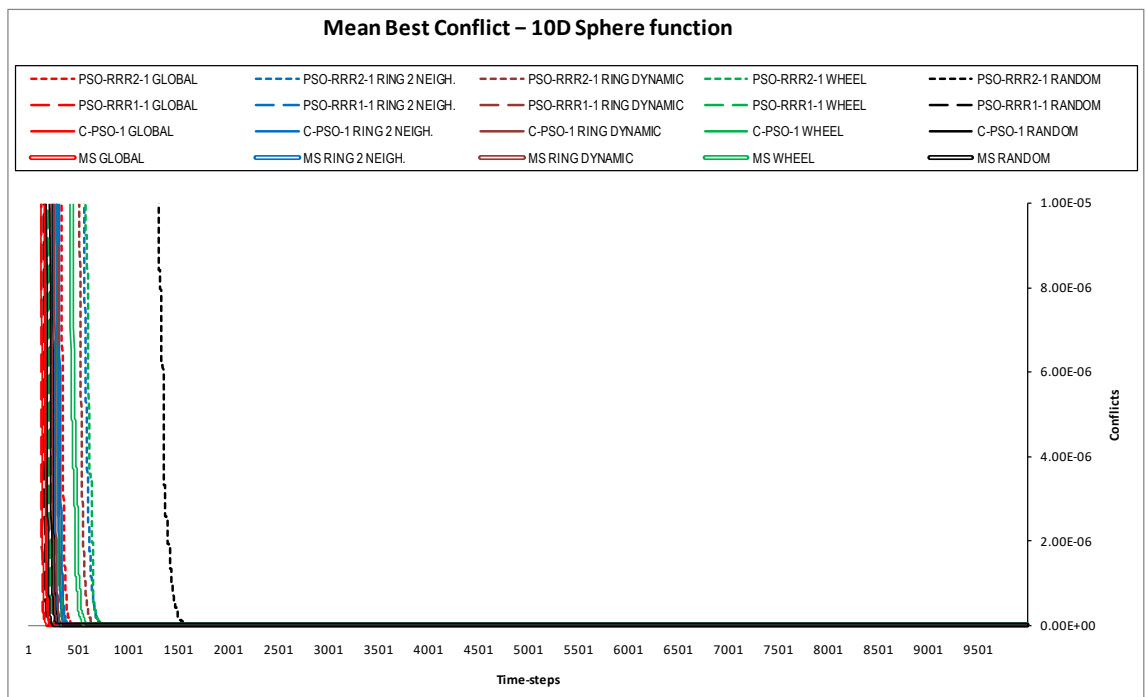

**Figure 3. Convergence curves of the mean best conflict for the 10D Sphere function, associated to Table 2. The colour-codes used to identify the neighbourhood structures are the same in the table and figure associated.**

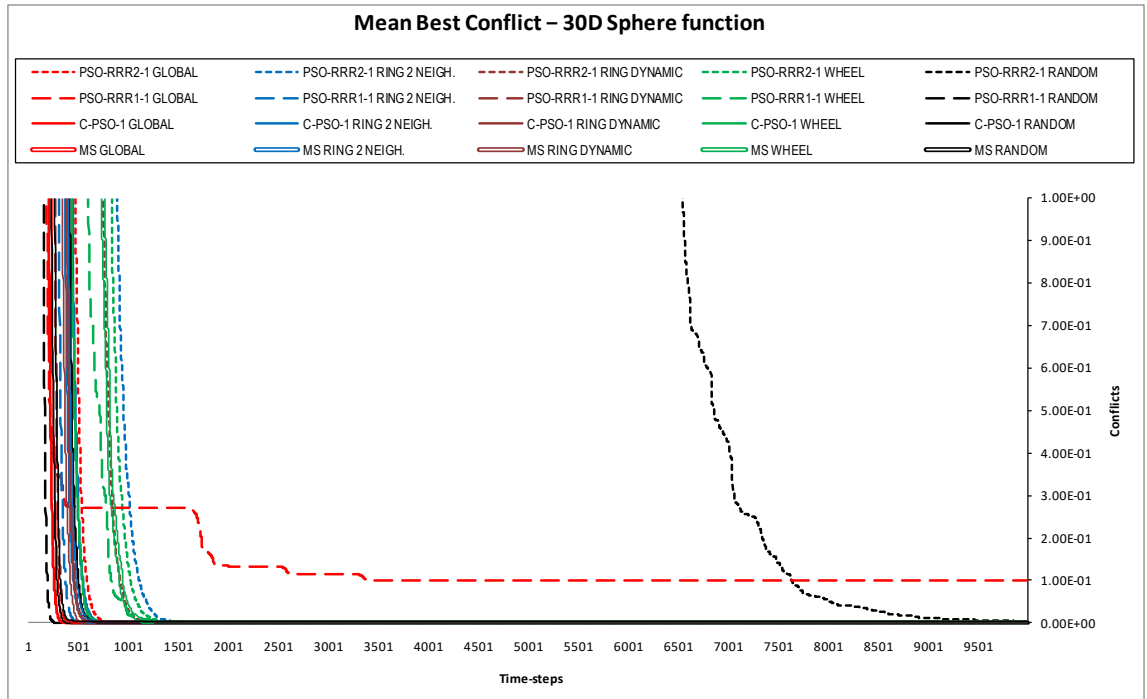

**Figure 4. Convergence curves of the mean best conflict for the 30D Sphere function, associated to Table 3. The colour-codes used to identify the neighbourhood structures are the same in the table and figure associated.**

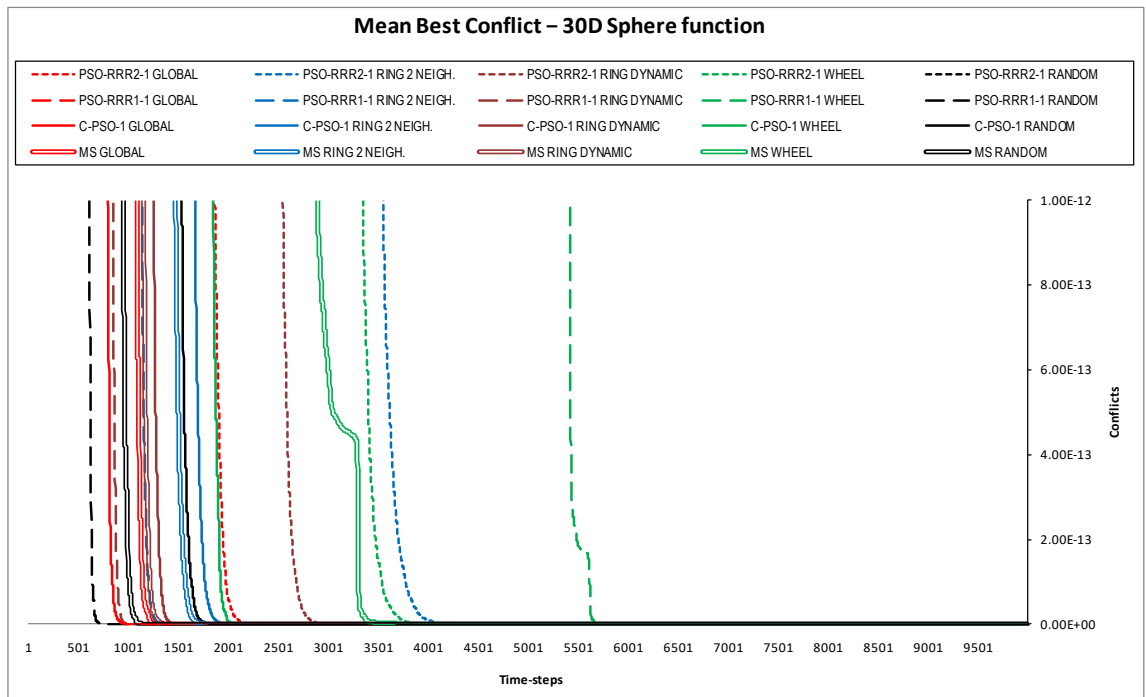

**Figure 5. Convergence curves of the mean best conflict for the 30D Sphere function, associated to Table 3. The colour-codes used to identify the neighbourhood structures are the same in the table and figure associated.**

**Table 4. Statistical results out of 25 runs for the PSO-RRR2-1, the PSO-RRR1-1, the C-PSO-1, and a Multi-Swarm algorithm optimizing the 2-dimensional Rosenbrock function. The neighbourhoods tested are the GLOBAL; the RING with 2 neighbours; the RING with linearly increasing number of neighbours (from 2 to 'swarm-size – 1'); the WHEEL; and a RANDOM topology. A run with an error no greater than 0.0001 is regarded as successful.**

| OPTIMIZER | NEIGHBOURHOOD STRUCTURE | | Time-steps | ROSENBROCK 2D | | | | OPTIMUM = 0 | |
|---|---|---|---|---|---|---|---|---|---|
| | | | | BEST | MEDIAN | MEAN | WORST | MEAN PB_ME | [%] Success |
| PSO-RRR2-1 | GLOBAL | | 10000 | 0.00E+00 | 0.00E+00 | 0.00E+00 | 0.00E+00 | 0.00E+00 | 100 |
| | | | 1000 | 1.54E-30 | 3.01E-28 | 4.82E-26 | 3.96E-25 | 3.72E-08 | - |
| | RING | nn = 2 | 10000 | 0.00E+00 | 0.00E+00 | 0.00E+00 | 0.00E+00 | 0.00E+00 | 100 |
| | | | 1000 | 3.11E-20 | 3.03E-16 | 3.31E-15 | 3.09E-14 | 6.95E-06 | - |
| | | nni = 2 nnf = (m – 1) | 10000 | 0.00E+00 | 0.00E+00 | 0.00E+00 | 0.00E+00 | 0.00E+00 | 100 |
| | | | 1000 | 3.71E-23 | 1.08E-18 | 2.30E-17 | 4.07E-16 | 4.46E-06 | - |
| | WHEEL | | 10000 | 0.00E+00 | 0.00E+00 | 0.00E+00 | 0.00E+00 | 0.00E+00 | 100 |
| | | | 1000 | 6.68E-25 | 1.00E-19 | 1.96E-15 | 4.47E-14 | 8.76E-07 | - |
| | RANDOM | | 10000 | 0.00E+00 | 0.00E+00 | 0.00E+00 | 0.00E+00 | 0.00E+00 | 100 |
| | | | 1000 | 4.13E-20 | 5.56E-17 | 2.58E-15 | 5.86E-14 | 7.67E-07 | - |
| PSO-RRR1-1 | GLOBAL | | 10000 | 0.00E+00 | 0.00E+00 | 0.00E+00 | 0.00E+00 | 0.00E+00 | 100 |
| | | | 1000 | 0.00E+00 | 0.00E+00 | 0.00E+00 | 0.00E+00 | 3.70E-20 | - |
| | RING | nn = 2 | 10000 | 0.00E+00 | 0.00E+00 | 0.00E+00 | 0.00E+00 | 0.00E+00 | 100 |
| | | | 1000 | 4.93E-32 | 4.78E-27 | 6.69E-25 | 7.34E-24 | 1.27E-11 | - |
| | | nni = 2 nnf = (m – 1) | 10000 | 0.00E+00 | 0.00E+00 | 0.00E+00 | 0.00E+00 | 0.00E+00 | 100 |
| | | | 1000 | 0.00E+00 | 0.00E+00 | 7.89E-33 | 1.97E-31 | 6.36E-15 | - |
| | WHEEL | | 10000 | 0.00E+00 | 0.00E+00 | 0.00E+00 | 0.00E+00 | 0.00E+00 | 100 |
| | | | 1000 | 0.00E+00 | 0.00E+00 | 3.20E-32 | 7.89E-31 | 3.08E-16 | - |
| | RANDOM | | 10000 | 0.00E+00 | 0.00E+00 | 0.00E+00 | 0.00E+00 | 0.00E+00 | 100 |
| | | | 1000 | 0.00E+00 | 0.00E+00 | 0.00E+00 | 0.00E+00 | 2.39E-18 | - |
| C-PSO-1 | GLOBAL | | 10000 | 0.00E+00 | 0.00E+00 | 0.00E+00 | 0.00E+00 | 0.00E+00 | 100 |
| | | | 1000 | 0.00E+00 | 0.00E+00 | 0.00E+00 | 0.00E+00 | 4.29E-11 | - |
| | RING | nn = 2 | 10000 | 0.00E+00 | 0.00E+00 | 0.00E+00 | 0.00E+00 | 0.00E+00 | 100 |
| | | | 1000 | 5.77E-21 | 1.65E-15 | 6.19E-14 | 1.30E-12 | 2.76E-06 | - |
| | | nni = 2 nnf = (m – 1) | 10000 | 0.00E+00 | 0.00E+00 | 0.00E+00 | 0.00E+00 | 0.00E+00 | 100 |
| | | | 1000 | 6.36E-24 | 2.02E-20 | 4.32E-19 | 5.48E-18 | 1.46E-07 | - |
| | WHEEL | | 10000 | 0.00E+00 | 0.00E+00 | 0.00E+00 | 0.00E+00 | 0.00E+00 | 100 |
| | | | 1000 | 4.44E-31 | 7.89E-26 | 2.30E-18 | 3.61E-17 | 6.35E-09 | - |
| | RANDOM | | 10000 | 0.00E+00 | 0.00E+00 | 0.00E+00 | 0.00E+00 | 0.00E+00 | 100 |
| | | | 1000 | 0.00E+00 | 1.11E-29 | 8.77E-28 | 1.56E-26 | 3.12E-09 | - |
| Multi-Swarm | GLOBAL | | 10000 | 0.00E+00 | 0.00E+00 | 0.00E+00 | 0.00E+00 | 0.00E+00 | 100 |
| | | | 1000 | 0.00E+00 | 0.00E+00 | 0.00E+00 | 0.00E+00 | 8.05E-09 | - |
| | RING | nn = 2 | 10000 | 0.00E+00 | 0.00E+00 | 0.00E+00 | 0.00E+00 | 0.00E+00 | 100 |
| | | | 1000 | 0.00E+00 | 9.00E-24 | 2.19E-18 | 3.50E-17 | 1.78E-06 | - |
| | | nni = 2 nnf = (m – 1) | 10000 | 0.00E+00 | 0.00E+00 | 0.00E+00 | 0.00E+00 | 0.00E+00 | 100 |
| | | | 1000 | 0.00E+00 | 1.77E-30 | 1.67E-26 | 3.95E-25 | 1.36E-06 | - |
| | WHEEL | | 10000 | 0.00E+00 | 0.00E+00 | 0.00E+00 | 0.00E+00 | 0.00E+00 | 100 |
| | | | 1000 | 2.61E-29 | 7.67E-24 | 1.99E-16 | 4.89E-15 | 4.17E-07 | - |
| | RANDOM | | 10000 | 0.00E+00 | 0.00E+00 | 0.00E+00 | 0.00E+00 | 0.00E+00 | 100 |
| | | | 1000 | 0.00E+00 | 0.00E+00 | 1.41E-28 | 3.49E-27 | 5.78E-09 | - |

**Table 5. Statistical results out of 25 runs for the PSO-RRR2-1, the PSO-RRR1-1, the C-PSO-1, and a Multi-Swarm algorithm optimizing the 10-dimensional Rosenbrock function. The neighbourhoods tested are the GLOBAL; the RING with 2 neighbours; the RING with linearly increasing number of neighbours (from 2 to 'swarm-size – 1'); the WHEEL; and a RANDOM topology. A run with an error no greater than 0.0001 is regarded as successful.**

| OPTIMIZER | NEIGHBOURHOOD STRUCTURE | | Time-steps | ROSENBROCK 10D | | | | OPTIMUM = 0 | |
|---|---|---|---|---|---|---|---|---|---|
| | | | | BEST | MEDIAN | MEAN | WORST | MEAN PB_ME | [%] Success |
| PSO-RRR2-1 | GLOBAL | | 10000 | 1.09E-06 | 2.72E-04 | 6.38E-01 | 3.99E+00 | 6.79E-03 | 32 |
| | | | 1000 | 2.22E-02 | 2.47E+00 | 5.31E+00 | 6.85E+01 | 2.34E-03 | - |
| | RING | nn = 2 | 10000 | 6.79E-05 | 1.64E-02 | 1.82E-02 | 7.14E-02 | 7.64E-03 | 4 |
| | | | 1000 | 9.88E-03 | 1.64E+00 | 2.03E+00 | 5.50E+00 | 1.63E-02 | - |
| | | nni = 2 nnf = (m − 1) | 10000 | 1.74E-06 | 3.60E-04 | 3.93E-04 | 1.15E-03 | 1.20E-02 | 24 |
| | | | 1000 | 1.72E-02 | 4.02E+00 | 3.24E+00 | 5.10E+00 | 1.71E-02 | - |
| | WHEEL | | 10000 | 7.23E-04 | 1.57E-03 | 3.21E-01 | 3.99E+00 | 4.79E-03 | 0 |
| | | | 1000 | 3.35E+00 | 4.59E+00 | 4.70E+00 | 7.49E+00 | 6.32E-04 | - |
| | RANDOM | | 10000 | 1.14E-01 | 1.57E+00 | 1.95E+00 | 5.93E+00 | 1.03E-03 | 0 |
| | | | 1000 | 6.91E-01 | 1.06E+01 | 3.96E+01 | 2.06E+02 | 8.98E-03 | - |
| PSO-RRR1-1 | GLOBAL | | 10000 | 1.13E-28 | 8.73E-01 | 1.64E+00 | 3.99E+00 | 1.08E-03 | 44 |
| | | | 1000 | 6.53E-06 | 1.46E+00 | 1.90E+00 | 4.99E+00 | 2.85E-03 | - |
| | RING | nn = 2 | 10000 | 2.20E-10 | 3.45E-09 | 3.21E-08 | 5.63E-07 | 2.39E-03 | 100 |
| | | | 1000 | 5.10E-04 | 1.17E+00 | 1.30E+00 | 4.19E+00 | 1.11E-02 | - |
| | | nni = 2 nnf = (m − 1) | 10000 | 5.92E-29 | 1.37E-28 | 1.59E-01 | 3.99E+00 | 1.36E-03 | 96 |
| | | | 1000 | 1.51E-04 | 1.16E-01 | 4.36E-01 | 4.17E+00 | 7.96E-03 | - |
| | WHEEL | | 10000 | 5.31E-22 | 1.26E-17 | 7.97E-01 | 3.99E+00 | 2.60E-04 | 80 |
| | | | 1000 | 9.02E-03 | 3.45E-01 | 1.11E+00 | 4.46E+00 | 6.77E-04 | - |
| | RANDOM | | 10000 | 1.44E-26 | 1.74E-22 | 3.19E-01 | 3.99E+00 | 2.59E-03 | 92 |
| | | | 1000 | 3.29E-03 | 9.70E-03 | 5.37E-01 | 4.98E+00 | 4.31E-03 | - |
| C-PSO-1 | GLOBAL | | 10000 | 1.18E-10 | 4.49E-06 | 4.79E-01 | 3.99E+00 | 8.56E-03 | 76 |
| | | | 1000 | 2.73E-03 | 5.03E-01 | 7.26E+00 | 8.06E+01 | 6.76E-03 | - |
| | RING | nn = 2 | 10000 | 1.23E-08 | 1.29E-03 | 1.61E-01 | 3.99E+00 | 7.32E-03 | 8 |
| | | | 1000 | 1.18E-02 | 3.08E+00 | 2.69E+00 | 5.08E+00 | 1.90E-02 | - |
| | | nni = 2 nnf = (m − 1) | 10000 | 3.97E-08 | 8.83E-07 | 6.94E-06 | 1.28E-04 | 8.35E-03 | 96 |
| | | | 1000 | 1.33E-03 | 2.28E+00 | 2.11E+00 | 3.79E+00 | 1.63E-02 | - |
| | WHEEL | | 10000 | 6.82E-07 | 1.55E-05 | 7.97E-01 | 3.99E+00 | 5.04E-03 | 72 |
| | | | 1000 | 4.74E-02 | 2.79E+00 | 8.92E+00 | 7.60E+01 | 1.54E-03 | - |
| | RANDOM | | 10000 | 2.90E-05 | 6.25E-04 | 4.79E-01 | 3.99E+00 | 8.31E-03 | 16 |
| | | | 1000 | 2.01E-01 | 2.51E+00 | 2.75E+00 | 6.57E+00 | 3.84E-03 | - |
| Multi-Swarm | GLOBAL | | 10000 | 5.91E-12 | 2.94E-02 | 1.02E+00 | 5.59E+00 | 7.53E-03 | 16 |
| | | | 1000 | 9.37E-05 | 6.67E-01 | 1.47E+00 | 6.46E+00 | 7.93E-03 | - |
| | RING | nn = 2 | 10000 | 4.01E-09 | 8.03E-07 | 1.61E-01 | 3.99E+00 | 8.87E-03 | 80 |
| | | | 1000 | 2.63E-03 | 1.62E+00 | 1.82E+00 | 5.11E+00 | 1.79E-02 | - |
| | | nni = 2 nnf = (m − 1) | 10000 | 1.88E-15 | 1.16E-09 | 1.42E-05 | 3.49E-04 | 1.08E-02 | 96 |
| | | | 1000 | 5.39E-04 | 6.18E-01 | 8.49E-01 | 4.07E+00 | 1.45E-02 | - |
| | WHEEL | | 10000 | 8.63E-07 | 1.79E-04 | 9.58E-01 | 4.00E+00 | 5.16E-03 | 32 |
| | | | 1000 | 2.41E-02 | 4.69E+00 | 5.62E+00 | 1.35E+01 | 3.19E-03 | - |
| | RANDOM | | 10000 | 6.03E-16 | 6.41E-15 | 6.38E-01 | 3.99E+00 | 5.68E-03 | 84 |
| | | | 1000 | 3.40E-02 | 2.42E-01 | 8.58E-01 | 4.27E+00 | 4.87E-03 | - |

**Table 6. Statistical results out of 25 runs for the PSO-RRR2-1, the PSO-RRR1-1, the C-PSO-1, and a Multi-Swarm algorithm optimizing the 30-dimensional Rosenbrock function. The neighbourhoods tested are the GLOBAL; the RING with 2 neighbours; the RING with linearly increasing number of neighbours (from 2 to 'swarm-size – 1'); the WHEEL; and a RANDOM topology. A run with an error no greater than 0.0001 is regarded as successful.**

| OPTIMIZER | NEIGHBOURHOOD STRUCTURE | | Time-steps | ROSENBROCK 30D | | | | OPTIMUM = 0 | |
|---|---|---|---|---|---|---|---|---|---|
| | | | | BEST | MEDIAN | MEAN | WORST | MEAN PB_ME | [%] Success |
| PSO-RRR2-1 | GLOBAL | | 10000 | 1.41E-04 | 1.27E+01 | 1.03E+01 | 1.88E+01 | 1.45E-03 | 0 |
| | | | 1000 | 8.48E+00 | 2.80E+01 | 5.20E+01 | 1.24E+02 | 2.95E-04 | - |
| | RING | nn = 2 | 10000 | 1.14E-01 | 1.00E+01 | 1.06E+01 | 2.31E+01 | 1.13E-02 | 0 |
| | | | 1000 | 4.82E+01 | 1.40E+02 | 1.48E+02 | 3.09E+02 | 1.57E-02 | - |
| | | nni = 2 nnf = (m – 1) | 10000 | 2.91E-07 | 1.46E+01 | 1.29E+01 | 2.28E+01 | 7.38E-03 | 4 |
| | | | 1000 | 2.84E+01 | 6.81E+01 | 7.44E+01 | 1.43E+02 | 1.16E-02 | - |
| | WHEEL | | 10000 | 3.17E+00 | 1.89E+01 | 2.63E+01 | 7.94E+01 | 6.73E-04 | 0 |
| | | | 1000 | 3.52E+01 | 1.11E+02 | 1.60E+02 | 5.59E+02 | 3.21E-04 | - |
| | RANDOM | | 10000 | 1.33E+01 | 9.97E+01 | 1.63E+02 | 5.42E+02 | 1.54E-03 | 0 |
| | | | 1000 | 8.75E+04 | 2.48E+05 | 3.90E+05 | 1.94E+06 | 2.56E-02 | - |
| PSO-RRR1-1 | GLOBAL | | 10000 | 2.27E+01 | 9.15E+01 | 1.06E+02 | 3.67E+02 | 3.98E-11 | 0 |
| | | | 1000 | 2.43E+01 | 9.38E+01 | 1.10E+02 | 3.69E+02 | 2.84E-08 | - |
| | RING | nn = 2 | 10000 | 8.78E-03 | 7.24E+00 | 7.16E+00 | 1.91E+01 | 2.15E-03 | 0 |
| | | | 1000 | 8.28E+00 | 2.61E+01 | 4.28E+01 | 1.77E+02 | 4.38E-03 | - |
| | | nni = 2 nnf = (m – 1) | 10000 | 2.56E-17 | 9.97E-13 | 1.35E+00 | 9.73E+00 | 2.25E-03 | 72 |
| | | | 1000 | 2.21E+01 | 2.48E+01 | 3.96E+01 | 8.33E+01 | 2.28E-03 | - |
| | WHEEL | | 10000 | 5.13E-04 | 2.33E+00 | 8.76E+00 | 8.10E+01 | 4.75E-06 | 0 |
| | | | 1000 | 2.84E+01 | 9.14E+01 | 1.16E+02 | 2.21E+02 | 7.36E-05 | - |
| | RANDOM | | 10000 | 2.12E-14 | 7.86E-10 | 1.12E+00 | 3.99E+00 | 6.76E-04 | 72 |
| | | | 1000 | 4.23E+00 | 2.08E+01 | 3.14E+01 | 1.22E+02 | 1.54E-04 | - |
| C-PSO-1 | GLOBAL | | 10000 | 1.17E-05 | 3.90E-02 | 1.05E+00 | 4.02E+00 | 5.58E-03 | 8 |
| | | | 1000 | 1.55E+00 | 2.22E+01 | 3.58E+01 | 1.79E+02 | 9.02E-04 | - |
| | RING | nn = 2 | 10000 | 2.89E-03 | 6.94E-01 | 3.39E+00 | 1.79E+01 | 7.42E-03 | 0 |
| | | | 1000 | 1.32E+01 | 2.89E+01 | 5.04E+01 | 1.45E+02 | 1.30E-02 | - |
| | | nni = 2 nnf = (m – 1) | 10000 | 3.41E-06 | 3.73E+00 | 3.16E+00 | 1.00E+01 | 6.55E-03 | 4 |
| | | | 1000 | 1.13E+01 | 2.63E+01 | 3.40E+01 | 8.12E+01 | 8.11E-03 | - |
| | WHEEL | | 10000 | 1.02E-03 | 4.82E+00 | 4.80E+00 | 1.07E+01 | 1.33E-03 | 0 |
| | | | 1000 | 2.04E+01 | 7.57E+01 | 6.24E+01 | 1.77E+02 | 2.16E-05 | - |
| | RANDOM | | 10000 | 3.07E-03 | 1.32E+01 | 1.09E+01 | 7.22E+01 | 1.63E-03 | 0 |
| | | | 1000 | 1.68E+01 | 6.96E+01 | 6.14E+01 | 2.16E+02 | 1.10E-04 | - |
| Multi-Swarm | GLOBAL | | 10000 | 4.21E-08 | 1.67E+01 | 2.70E+01 | 7.68E+01 | 3.90E-08 | 4 |
| | | | 1000 | 2.33E-02 | 2.30E+01 | 4.38E+01 | 1.36E+02 | 2.59E-03 | - |
| | RING | nn = 2 | 10000 | 9.14E-03 | 7.09E+00 | 6.59E+00 | 1.46E+01 | 3.01E-03 | 0 |
| | | | 1000 | 6.26E+00 | 7.08E+01 | 5.33E+01 | 8.71E+01 | 5.27E-03 | - |
| | | nni = 2 nnf = (m – 1) | 10000 | 1.56E-05 | 5.30E+00 | 6.36E+00 | 1.98E+01 | 2.76E-03 | 4 |
| | | | 1000 | 4.72E+00 | 2.73E+01 | 4.05E+01 | 1.34E+02 | 3.79E-03 | - |
| | WHEEL | | 10000 | 2.16E-02 | 1.33E+01 | 1.98E+01 | 7.13E+01 | 6.47E-04 | 0 |
| | | | 1000 | 2.67E+01 | 1.29E+02 | 1.30E+02 | 2.98E+02 | 1.17E-04 | - |
| | RANDOM | | 10000 | 2.60E-04 | 1.16E+00 | 2.11E+00 | 1.15E+01 | 5.07E-03 | 0 |
| | | | 1000 | 7.09E+00 | 2.55E+01 | 4.35E+01 | 1.14E+02 | 2.25E-03 | - |

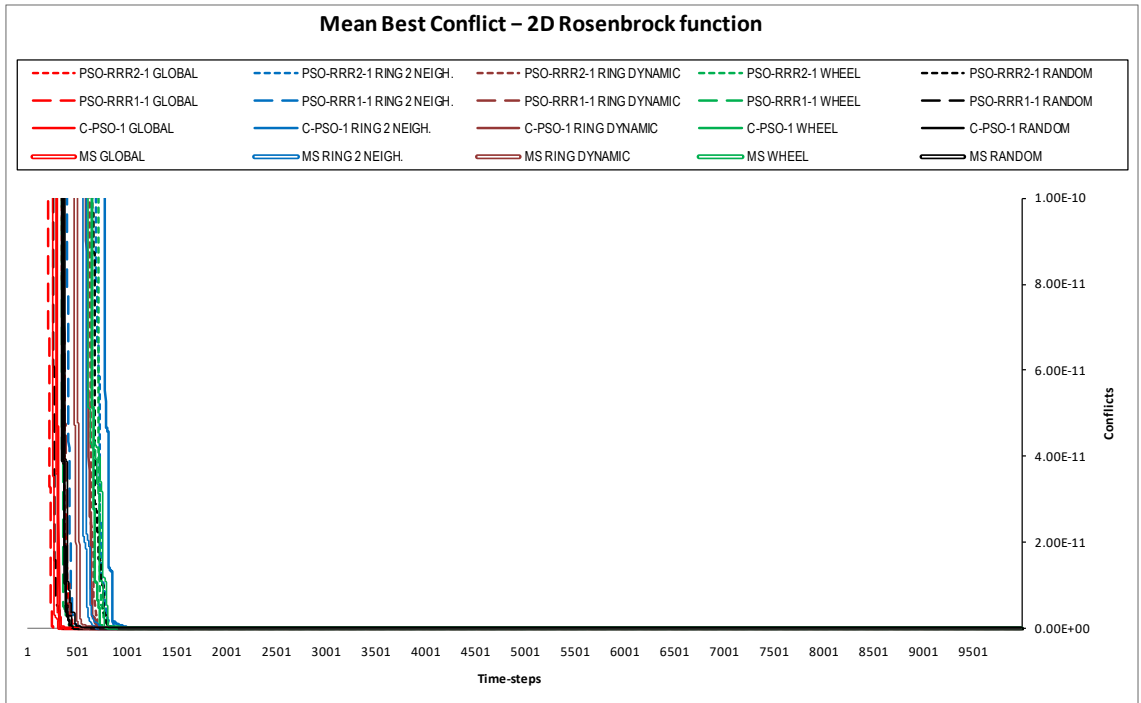

**Figure 6. Convergence curves of the mean best conflict for the 2D Rosenbrock function, associated to Table 4. The colour-codes used to identify the neighbourhood structures are the same in the table and figure associated.**

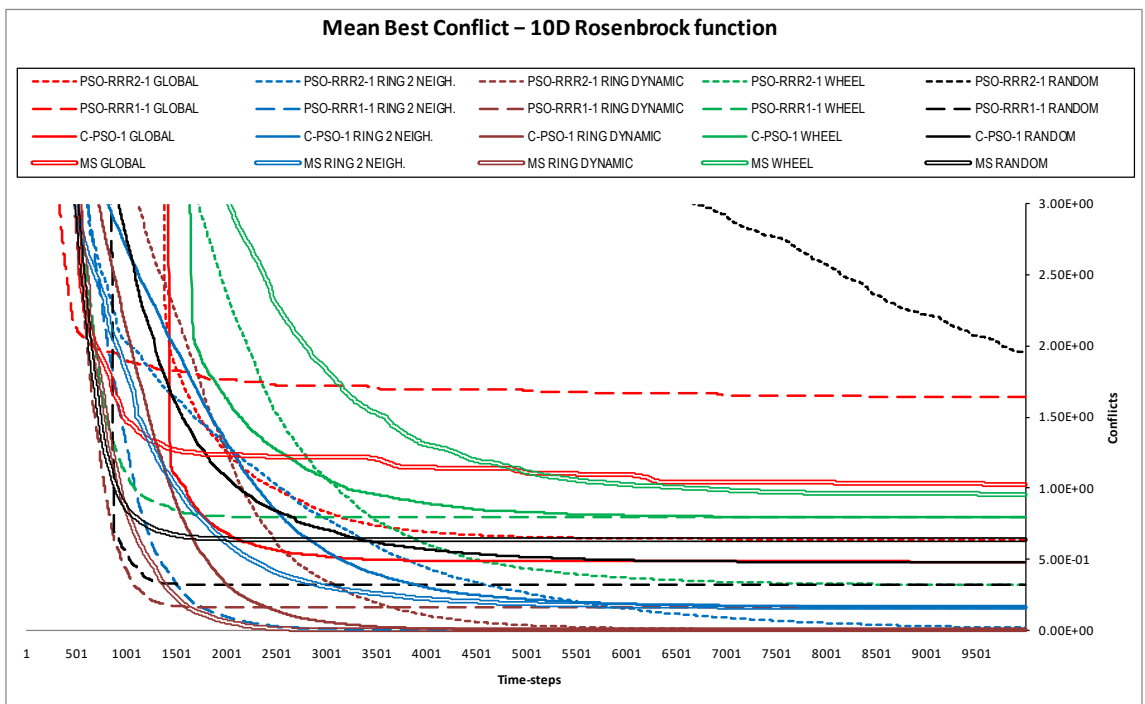

**Figure 7. Convergence curves of the mean best conflict for the 10D Rosenbrock function, associated to Table 5. The colour-codes used to identify the neighbourhood structures are the same in the table and figure associated.**

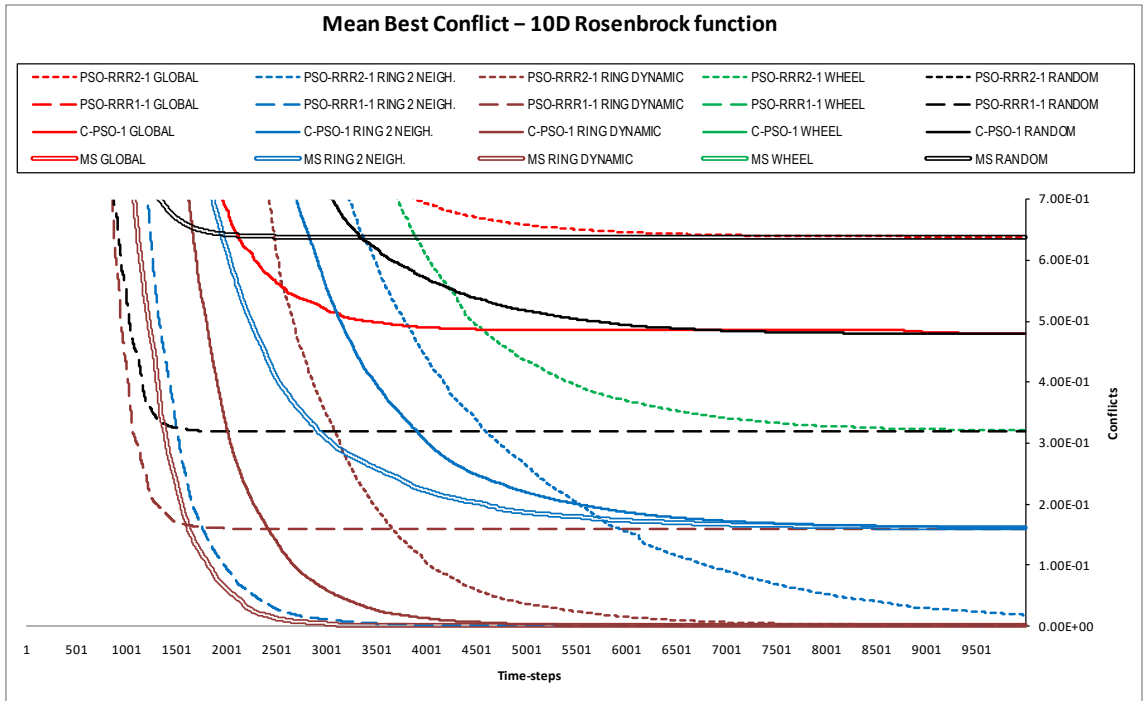

**Figure 8. Convergence curves of the mean best conflict for the 10D Rosenbrock function, associated to Table 5. The colour-codes used to identify the neighbourhood structures are the same in the table and figure associated.**

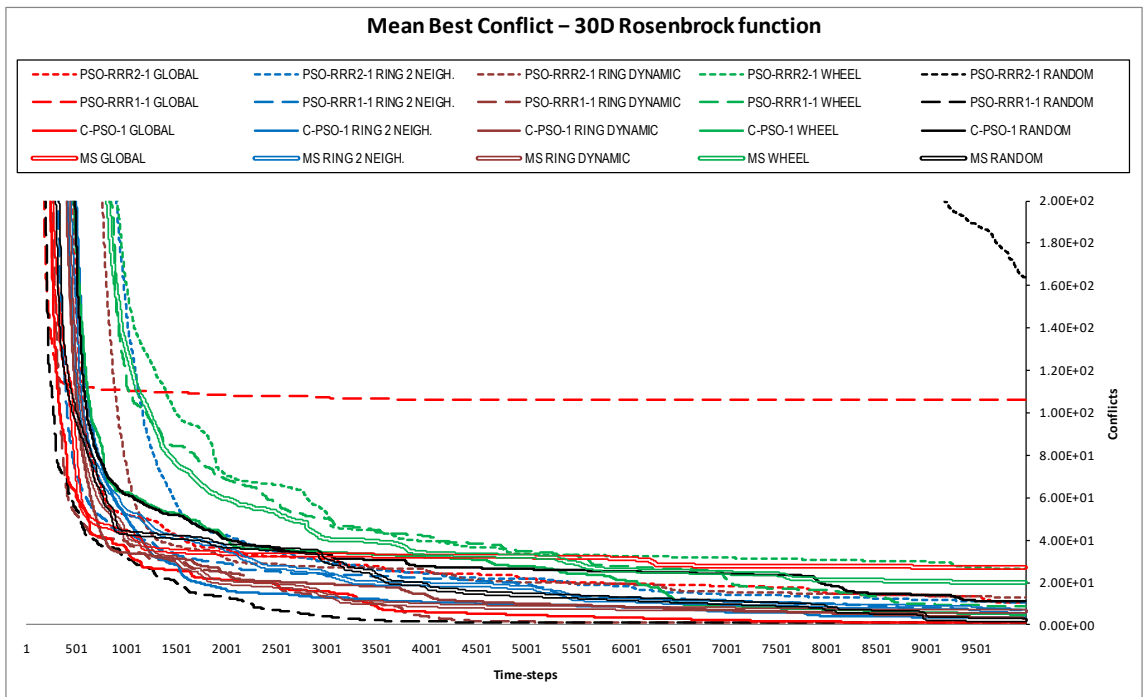

**Figure 9. Convergence curves of the mean best conflict for the 30D Rosenbrock function, associated to Table 6. The colour-codes used to identify the neighbourhood structures are the same in the table and figure associated.**

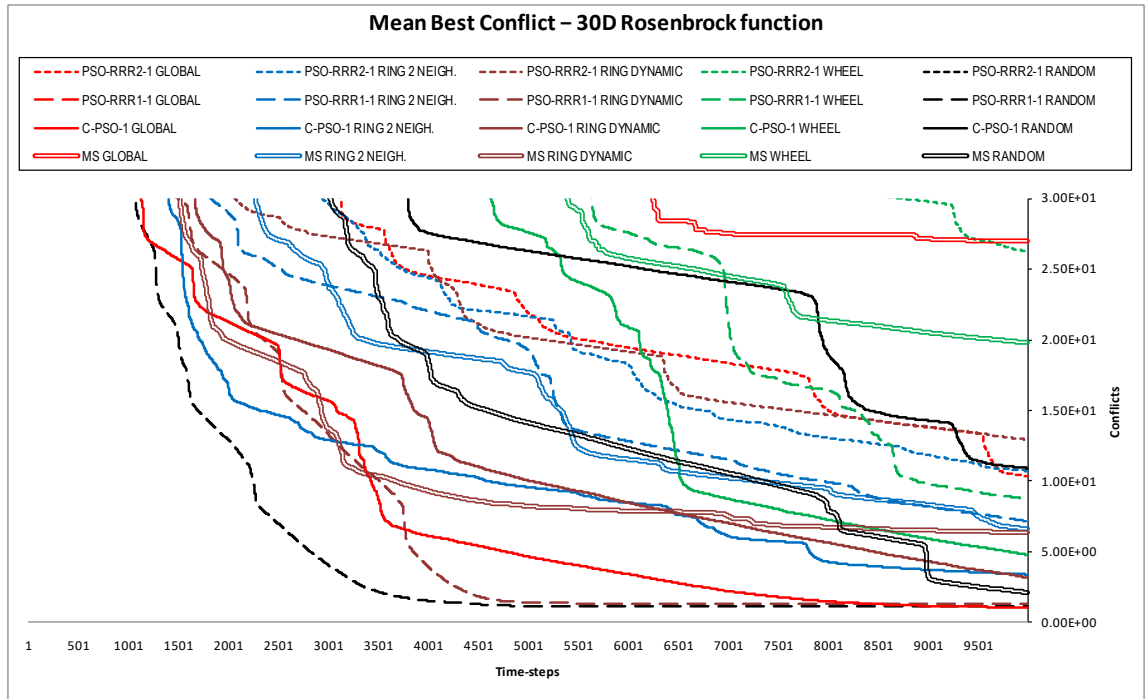

**Figure 10. Convergence curves of the mean best conflict for the 30D Rosenbrock function, associated to Table 6. The colour-codes used to identify the neighbourhood structures are the same in the table and figure associated.**

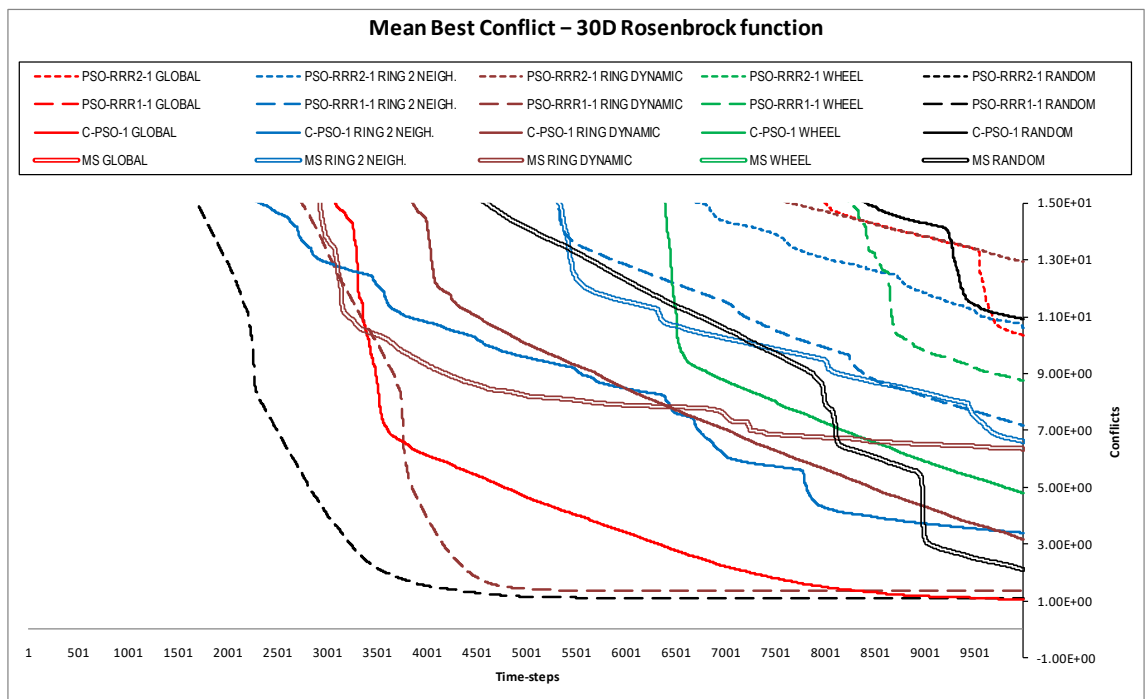

**Figure 11. Convergence curves of the mean best conflict for the 30D Rosenbrock function, associated to Table 6. The colour-codes used to identify the neighbourhood structures are the same in the table and figure associated.**

**Table 7. Statistical results out of 25 runs for the PSO-RRR2-1, the PSO-RRR1-1, the C-PSO-1, and a Multi-Swarm algorithm optimizing the 2-dimensional Rastrigin function. The neighbourhoods tested are the GLOBAL; the RING with 2 neighbours; the RING with linearly increasing number of neighbours (from 2 to 'swarm-size – 1'); the WHEEL; and a RANDOM topology. A run with an error no greater than 0.0001 is regarded as successful.**

| OPTIMIZER | NEIGHBOURHOOD STRUCTURE | | Time-steps | RASTRIGIN 2D | | | | OPTIMUM = 0 | |
|---|---|---|---|---|---|---|---|---|---|
| | | | | BEST | MEDIAN | MEAN | WORST | MEAN PB_ME | [%] Success |
| PSO-RRR2-1 | GLOBAL | | 10000 | 0.00E+00 | 0.00E+00 | 0.00E+00 | 0.00E+00 | 1.07E-10 | 100 |
| | | | 1000 | 0.00E+00 | 0.00E+00 | 0.00E+00 | 0.00E+00 | 9.42E-11 | - |
| | RING | nn = 2 | 10000 | 0.00E+00 | 0.00E+00 | 0.00E+00 | 0.00E+00 | 9.93E-11 | 100 |
| | | | 1000 | 0.00E+00 | 0.00E+00 | 0.00E+00 | 0.00E+00 | 9.93E-11 | - |
| | | nni = 2 nnf = (m – 1) | 10000 | 0.00E+00 | 0.00E+00 | 0.00E+00 | 0.00E+00 | 8.96E-11 | 100 |
| | | | 1000 | 0.00E+00 | 0.00E+00 | 0.00E+00 | 0.00E+00 | 8.96E-11 | - |
| | WHEEL | | 10000 | 0.00E+00 | 0.00E+00 | 0.00E+00 | 0.00E+00 | 9.01E-11 | 100 |
| | | | 1000 | 0.00E+00 | 0.00E+00 | 0.00E+00 | 0.00E+00 | 9.01E-11 | - |
| | RANDOM | | 10000 | 0.00E+00 | 0.00E+00 | 0.00E+00 | 0.00E+00 | 9.33E-11 | 100 |
| | | | 1000 | 0.00E+00 | 0.00E+00 | 0.00E+00 | 0.00E+00 | 9.33E-11 | - |
| PSO-RRR1-1 | GLOBAL | | 10000 | 0.00E+00 | 0.00E+00 | 0.00E+00 | 0.00E+00 | 8.09E-11 | 100 |
| | | | 1000 | 0.00E+00 | 0.00E+00 | 0.00E+00 | 0.00E+00 | 8.09E-11 | - |
| | RING | nn = 2 | 10000 | 0.00E+00 | 0.00E+00 | 0.00E+00 | 0.00E+00 | 9.05E-11 | 100 |
| | | | 1000 | 0.00E+00 | 0.00E+00 | 0.00E+00 | 0.00E+00 | 9.05E-11 | - |
| | | nni = 2 nnf = (m – 1) | 10000 | 0.00E+00 | 0.00E+00 | 0.00E+00 | 0.00E+00 | 8.60E-11 | 100 |
| | | | 1000 | 0.00E+00 | 0.00E+00 | 0.00E+00 | 0.00E+00 | 8.60E-11 | - |
| | WHEEL | | 10000 | 0.00E+00 | 0.00E+00 | 0.00E+00 | 0.00E+00 | 8.45E-11 | 100 |
| | | | 1000 | 0.00E+00 | 0.00E+00 | 0.00E+00 | 0.00E+00 | 8.45E-11 | - |
| | RANDOM | | 10000 | 0.00E+00 | 0.00E+00 | 0.00E+00 | 0.00E+00 | 9.35E-11 | 100 |
| | | | 1000 | 0.00E+00 | 0.00E+00 | 0.00E+00 | 0.00E+00 | 9.35E-11 | - |
| C-PSO-1 | GLOBAL | | 10000 | 0.00E+00 | 0.00E+00 | 0.00E+00 | 0.00E+00 | 7.79E-11 | 100 |
| | | | 1000 | 0.00E+00 | 0.00E+00 | 0.00E+00 | 0.00E+00 | 7.79E-11 | - |
| | RING | nn = 2 | 10000 | 0.00E+00 | 0.00E+00 | 0.00E+00 | 0.00E+00 | 8.52E-11 | 100 |
| | | | 1000 | 0.00E+00 | 0.00E+00 | 0.00E+00 | 0.00E+00 | 8.52E-11 | - |
| | | nni = 2 nnf = (m – 1) | 10000 | 0.00E+00 | 0.00E+00 | 0.00E+00 | 0.00E+00 | 9.26E-11 | 100 |
| | | | 1000 | 0.00E+00 | 0.00E+00 | 0.00E+00 | 0.00E+00 | 9.26E-11 | - |
| | WHEEL | | 10000 | 0.00E+00 | 0.00E+00 | 0.00E+00 | 0.00E+00 | 9.03E-11 | 100 |
| | | | 1000 | 0.00E+00 | 0.00E+00 | 0.00E+00 | 0.00E+00 | 9.03E-11 | - |
| | RANDOM | | 10000 | 0.00E+00 | 0.00E+00 | 0.00E+00 | 0.00E+00 | 9.07E-11 | 100 |
| | | | 1000 | 0.00E+00 | 0.00E+00 | 0.00E+00 | 0.00E+00 | 9.07E-11 | - |
| Multi-Swarm | GLOBAL | | 10000 | 0.00E+00 | 0.00E+00 | 0.00E+00 | 0.00E+00 | 9.93E-11 | 100 |
| | | | 1000 | 0.00E+00 | 0.00E+00 | 0.00E+00 | 0.00E+00 | 9.93E-11 | - |
| | RING | nn = 2 | 10000 | 0.00E+00 | 0.00E+00 | 0.00E+00 | 0.00E+00 | 8.73E-11 | 100 |
| | | | 1000 | 2.98E+00 | 5.97E+00 | 5.72E+00 | 1.00E+01 | 3.74E-02 | - |
| | | nni = 2 nnf = (m – 1) | 10000 | 0.00E+00 | 0.00E+00 | 0.00E+00 | 0.00E+00 | 9.16E-11 | 100 |
| | | | 1000 | 0.00E+00 | 0.00E+00 | 0.00E+00 | 0.00E+00 | 9.16E-11 | - |
| | WHEEL | | 10000 | 0.00E+00 | 0.00E+00 | 0.00E+00 | 0.00E+00 | 8.81E-11 | 100 |
| | | | 1000 | 0.00E+00 | 0.00E+00 | 0.00E+00 | 0.00E+00 | 8.81E-11 | - |
| | RANDOM | | 10000 | 0.00E+00 | 0.00E+00 | 0.00E+00 | 0.00E+00 | 9.49E-11 | 100 |
| | | | 1000 | 0.00E+00 | 0.00E+00 | 0.00E+00 | 0.00E+00 | 9.49E-11 | - |

**Table 8. Statistical results out of 25 runs for the PSO-RRR2-1, the PSO-RRR1-1, the C-PSO-1, and a Multi-Swarm algorithm optimizing the 10-dimensional Rastrigin function. The neighbourhoods tested are the GLOBAL; the RING with 2 neighbours; the RING with linearly increasing number of neighbours (from 2 to 'swarm-size – 1'); the WHEEL; and a RANDOM topology. A run with an error no greater than 0.0001 is regarded as successful.**

| OPTIMIZER | NEIGHBOURHOOD STRUCTURE | | Time-steps | RASTRIGIN 10D | | | | OPTIMUM = 0 | |
|---|---|---|---|---|---|---|---|---|---|
| | | | | BEST | MEDIAN | MEAN | WORST | MEAN PB_ME | [%] Success |
| PSO-RRR2-1 | GLOBAL | | 10000 | 9.95E-01 | 2.98E+00 | 2.95E+00 | 6.96E+00 | 3.48E-04 | 0 |
| | | | 1000 | 9.95E-01 | 2.98E+00 | 3.02E+00 | 6.96E+00 | 2.20E-03 | - |
| | RING | nn = 2 | 10000 | 0.00E+00 | 1.99E+00 | 2.15E+00 | 4.97E+00 | 2.71E-02 | 20 |
| | | | 1000 | 1.99E+00 | 4.22E+00 | 4.43E+00 | 7.96E+00 | 3.54E-02 | - |
| | | nni = 2 nnf = (m – 1) | 10000 | 0.00E+00 | 0.00E+00 | 7.16E-01 | 3.98E+00 | 1.65E-02 | 60 |
| | | | 1000 | 9.95E-01 | 3.98E+00 | 4.04E+00 | 1.09E+01 | 3.51E-02 | - |
| | WHEEL | | 10000 | 9.95E-01 | 1.99E+00 | 2.47E+00 | 5.97E+00 | 8.48E-04 | 0 |
| | | | 1000 | 9.97E-01 | 2.99E+00 | 3.09E+00 | 5.97E+00 | 7.73E-03 | - |
| | RANDOM | | 10000 | 0.00E+00 | 1.19E+01 | 1.18E+01 | 2.87E+01 | 3.01E-02 | 12 |
| | | | 1000 | 7.30E+00 | 2.51E+01 | 2.45E+01 | 3.65E+01 | 5.00E-02 | - |
| PSO-RRR1-1 | GLOBAL | | 10000 | 5.97E+00 | 1.19E+01 | 1.35E+01 | 2.49E+01 | 1.21E-11 | 0 |
| | | | 1000 | 5.97E+00 | 1.19E+01 | 1.35E+01 | 2.49E+01 | 1.47E-11 | - |
| | RING | nn = 2 | 10000 | 9.95E-01 | 4.97E+00 | 5.18E+00 | 1.09E+01 | 3.82E-02 | 0 |
| | | | 1000 | 2.98E+00 | 7.96E+00 | 7.94E+00 | 1.37E+01 | 4.30E-02 | - |
| | | nni = 2 nnf = (m – 1) | 10000 | 0.00E+00 | 2.98E+00 | 3.02E+00 | 5.97E+00 | 2.79E-02 | 8 |
| | | | 1000 | 2.98E+00 | 5.97E+00 | 7.44E+00 | 1.59E+01 | 4.40E-02 | - |
| | WHEEL | | 10000 | 4.97E+00 | 8.95E+00 | 9.47E+00 | 1.99E+01 | 1.80E-04 | 0 |
| | | | 1000 | 4.97E+00 | 8.95E+00 | 9.83E+00 | 1.99E+01 | 1.91E-03 | - |
| | RANDOM | | 10000 | 9.95E-01 | 4.97E+00 | 5.21E+00 | 8.95E+00 | 2.78E-11 | 0 |
| | | | 1000 | 9.95E-01 | 6.96E+00 | 7.01E+00 | 2.63E+01 | 4.96E-03 | - |
| C-PSO-1 | GLOBAL | | 10000 | 1.99E+00 | 3.98E+00 | 4.93E+00 | 1.09E+01 | 1.92E-11 | 0 |
| | | | 1000 | 1.99E+00 | 4.97E+00 | 5.17E+00 | 1.19E+01 | 8.71E-04 | - |
| | RING | nn = 2 | 10000 | 0.00E+00 | 2.98E+00 | 2.79E+00 | 4.97E+00 | 2.93E-02 | 12 |
| | | | 1000 | 1.99E+00 | 3.98E+00 | 4.55E+00 | 7.96E+00 | 3.79E-02 | - |
| | | nni = 2 nnf = (m – 1) | 10000 | 0.00E+00 | 9.95E-01 | 1.23E+00 | 5.97E+00 | 1.98E-02 | 48 |
| | | | 1000 | 1.99E+00 | 3.98E+00 | 4.66E+00 | 7.96E+00 | 3.65E-02 | - |
| | WHEEL | | 10000 | 0.00E+00 | 2.98E+00 | 3.26E+00 | 7.96E+00 | 4.96E-04 | 4 |
| | | | 1000 | 0.00E+00 | 2.99E+00 | 3.54E+00 | 7.96E+00 | 2.41E-03 | - |
| | RANDOM | | 10000 | 0.00E+00 | 1.99E+00 | 2.71E+00 | 2.21E+01 | 2.64E-03 | 12 |
| | | | 1000 | 9.95E-01 | 5.97E+00 | 1.12E+01 | 3.50E+01 | 2.43E-02 | - |
| Multi-Swarm | GLOBAL | | 10000 | 1.99E+00 | 3.98E+00 | 4.78E+00 | 1.49E+01 | 1.84E-11 | 0 |
| | | | 1000 | 1.99E+00 | 3.98E+00 | 5.13E+00 | 1.49E+01 | 5.46E-04 | - |
| | RING | nn = 2 | 10000 | 0.00E+00 | 2.98E+00 | 2.75E+00 | 6.96E+00 | 2.90E-02 | 4 |
| | | | 1000 | 2.98E+00 | 5.97E+00 | 5.72E+00 | 1.00E+01 | 3.74E-02 | - |
| | | nni = 2 nnf = (m – 1) | 10000 | 0.00E+00 | 9.95E-01 | 1.68E+00 | 5.97E+00 | 1.94E-02 | 32 |
| | | | 1000 | 2.18E+00 | 4.97E+00 | 5.26E+00 | 1.09E+01 | 3.66E-02 | - |
| | WHEEL | | 10000 | 0.00E+00 | 3.98E+00 | 4.42E+00 | 8.95E+00 | 7.48E-04 | 4 |
| | | | 1000 | 2.42E-12 | 4.97E+00 | 4.82E+00 | 8.95E+00 | 4.10E-03 | - |
| | RANDOM | | 10000 | 9.95E-01 | 2.98E+00 | 2.91E+00 | 4.97E+00 | 1.93E-03 | 0 |
| | | | 1000 | 9.95E-01 | 4.98E+00 | 9.76E+00 | 3.80E+01 | 2.37E-02 | - |

**Table 9. Statistical results out of 25 runs for the PSO-RRR2-1, the PSO-RRR1-1, the C-PSO-1, and a Multi-Swarm algorithm optimizing the 30-dimensional Rastrigin function. The neighbourhoods tested are the GLOBAL; the RING with 2 neighbours; the RING with linearly increasing number of neighbours (from 2 to 'swarm-size – 1'); the WHEEL; and a RANDOM topology. A run with an error no greater than 0.0001 is regarded as successful.**

| OPTIMIZER | NEIGHBOURHOOD STRUCTURE | | Time-steps | RASTRIGIN 30D | | | | OPTIMUM = 0 | |
|---|---|---|---|---|---|---|---|---|---|
| | | | | BEST | MEDIAN | MEAN | WORST | MEAN PB_ME | [%] Success |
| PSO-RRR2-1 | GLOBAL | | 10000 | 2.69E+01 | 4.28E+01 | 4.13E+01 | 5.57E+01 | 2.64E-11 | 0 |
| | | | 1000 | 2.69E+01 | 4.28E+01 | 4.14E+01 | 5.57E+01 | 3.99E-05 | - |
| | RING | nn = 2 | 10000 | 2.98E+01 | 4.40E+01 | 4.29E+01 | 5.29E+01 | 2.61E-02 | 0 |
| | | | 1000 | 3.46E+01 | 5.32E+01 | 5.24E+01 | 7.23E+01 | 3.00E-02 | - |
| | | nni = 2 nnf = (m – 1) | 10000 | 2.69E+01 | 4.28E+01 | 4.31E+01 | 6.96E+01 | 2.97E-02 | 0 |
| | | | 1000 | 3.84E+01 | 4.88E+01 | 5.19E+01 | 7.99E+01 | 2.99E-02 | - |
| | WHEEL | | 10000 | 2.09E+01 | 3.68E+01 | 3.77E+01 | 6.77E+01 | 2.23E-04 | 0 |
| | | | 1000 | 2.82E+01 | 4.41E+01 | 4.54E+01 | 6.98E+01 | 5.67E-03 | - |
| | RANDOM | | 10000 | 2.50E+01 | 1.77E+02 | 1.42E+02 | 2.19E+02 | 2.81E-02 | 0 |
| | | | 1000 | 1.23E+02 | 2.01E+02 | 1.99E+02 | 2.48E+02 | 3.14E-02 | - |
| PSO-RRR1-1 | GLOBAL | | 10000 | 2.49E+01 | 7.16E+01 | 7.41E+01 | 1.28E+02 | 6.68E-16 | 0 |
| | | | 1000 | 2.49E+01 | 7.16E+01 | 7.41E+01 | 1.28E+02 | 1.47E-15 | - |
| | RING | nn = 2 | 10000 | 2.19E+01 | 4.68E+01 | 4.65E+01 | 6.17E+01 | 3.00E-02 | 0 |
| | | | 1000 | 2.20E+01 | 5.01E+01 | 5.03E+01 | 6.71E+01 | 3.12E-02 | - |
| | | nni = 2 nnf = (m – 1) | 10000 | 2.29E+01 | 4.88E+01 | 4.91E+01 | 7.46E+01 | 3.15E-02 | 0 |
| | | | 1000 | 3.48E+01 | 4.88E+01 | 5.13E+01 | 7.98E+01 | 3.27E-02 | - |
| | WHEEL | | 10000 | 4.58E+01 | 7.36E+01 | 6.90E+01 | 9.35E+01 | 4.60E-12 | 0 |
| | | | 1000 | 4.58E+01 | 7.36E+01 | 6.90E+01 | 9.35E+01 | 6.62E-04 | - |
| | RANDOM | | 10000 | 2.49E+01 | 4.78E+01 | 4.94E+01 | 7.46E+01 | 2.99E-11 | 0 |
| | | | 1000 | 2.49E+01 | 4.88E+01 | 5.98E+01 | 2.10E+02 | 1.58E-03 | - |
| C-PSO-1 | GLOBAL | | 10000 | 2.69E+01 | 4.88E+01 | 5.37E+01 | 9.65E+01 | 1.93E-11 | 0 |
| | | | 1000 | 2.69E+01 | 4.88E+01 | 5.37E+01 | 9.65E+01 | 1.09E-10 | - |
| | RING | nn = 2 | 10000 | 2.89E+01 | 5.37E+01 | 5.05E+01 | 6.87E+01 | 3.31E-02 | 0 |
| | | | 1000 | 2.89E+01 | 5.88E+01 | 5.59E+01 | 7.79E+01 | 3.43E-02 | - |
| | | nni = 2 nnf = (m – 1) | 10000 | 2.19E+01 | 5.27E+01 | 5.11E+01 | 7.36E+01 | 3.15E-02 | 0 |
| | | | 1000 | 2.69E+01 | 5.77E+01 | 5.59E+01 | 8.28E+01 | 3.36E-02 | - |
| | WHEEL | | 10000 | 2.98E+01 | 4.88E+01 | 5.13E+01 | 8.36E+01 | 3.01E-04 | 0 |
| | | | 1000 | 2.98E+01 | 4.88E+01 | 5.27E+01 | 8.36E+01 | 6.88E-04 | - |
| | RANDOM | | 10000 | 2.29E+01 | 3.48E+01 | 3.73E+01 | 6.57E+01 | 1.07E-04 | 0 |
| | | | 1000 | 2.30E+01 | 1.36E+02 | 1.28E+02 | 2.25E+02 | 1.91E-02 | - |
| Multi-Swarm | GLOBAL | | 10000 | 2.59E+01 | 5.27E+01 | 5.33E+01 | 8.16E+01 | 1.89E-11 | 0 |
| | | | 1000 | 2.59E+01 | 5.27E+01 | 5.33E+01 | 8.16E+01 | 5.36E-08 | - |
| | RING | nn = 2 | 10000 | 3.28E+01 | 4.48E+01 | 4.56E+01 | 6.37E+01 | 3.03E-02 | 0 |
| | | | 1000 | 3.32E+01 | 5.21E+01 | 4.97E+01 | 6.57E+01 | 3.09E-02 | - |
| | | nni = 2 nnf = (m – 1) | 10000 | 2.59E+01 | 3.98E+01 | 4.32E+01 | 6.67E+01 | 2.77E-02 | 0 |
| | | | 1000 | 2.72E+01 | 4.48E+01 | 4.77E+01 | 6.83E+01 | 2.96E-02 | - |
| | WHEEL | | 10000 | 1.69E+01 | 4.48E+01 | 4.61E+01 | 6.96E+01 | 1.04E-04 | 0 |
| | | | 1000 | 1.71E+01 | 4.68E+01 | 4.77E+01 | 7.77E+01 | 2.40E-03 | - |
| | RANDOM | | 10000 | 1.99E+01 | 3.78E+01 | 3.86E+01 | 6.67E+01 | 3.08E-11 | 0 |
| | | | 1000 | 2.19E+01 | 4.18E+01 | 6.60E+01 | 2.00E+02 | 5.51E-03 | - |

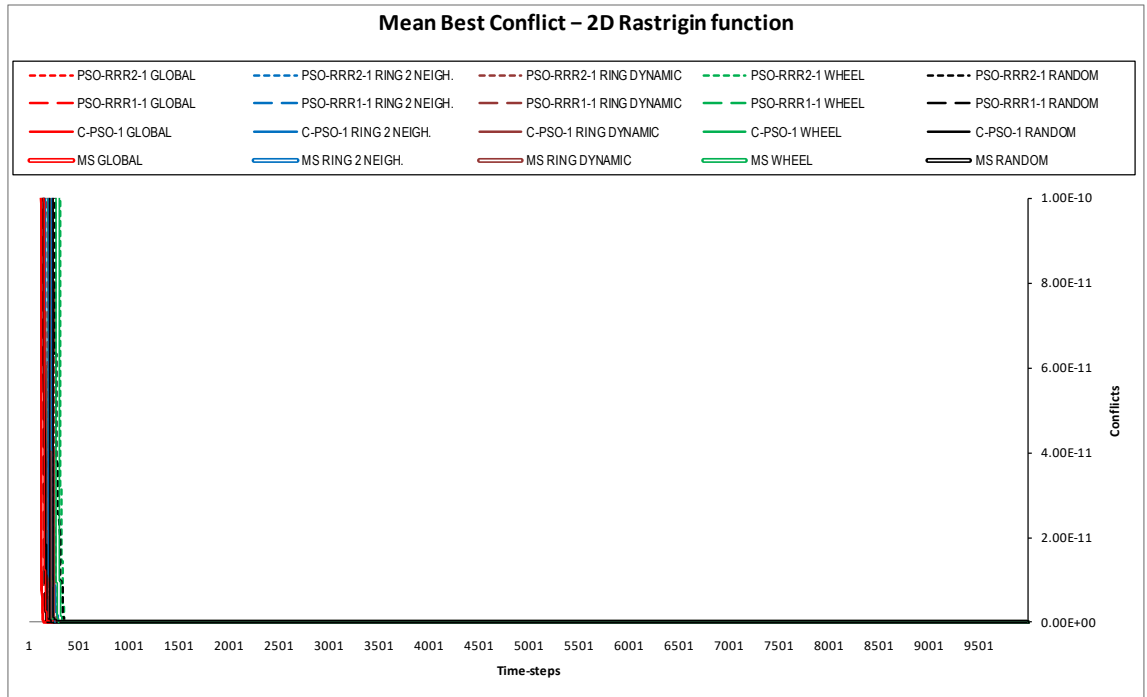

**Figure 12. Convergence curves of the mean best conflict for the 2D Rastrigin function, associated to Table 7. The colour-codes used to identify the neighbourhood structures are the same in the table and figure associated.**

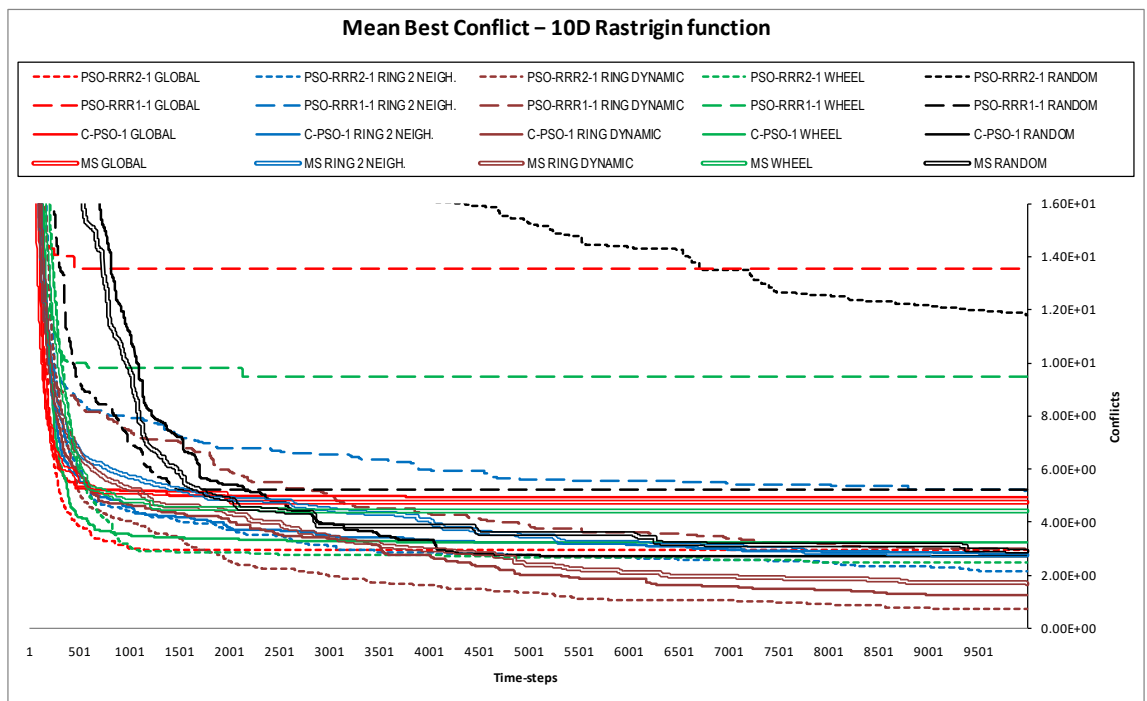

**Figure 13. Convergence curves of the mean best conflict for the 10D Rastrigin function, associated to Table 8. The colour-codes used to identify the neighbourhood structures are the same in the table and figure associated.**

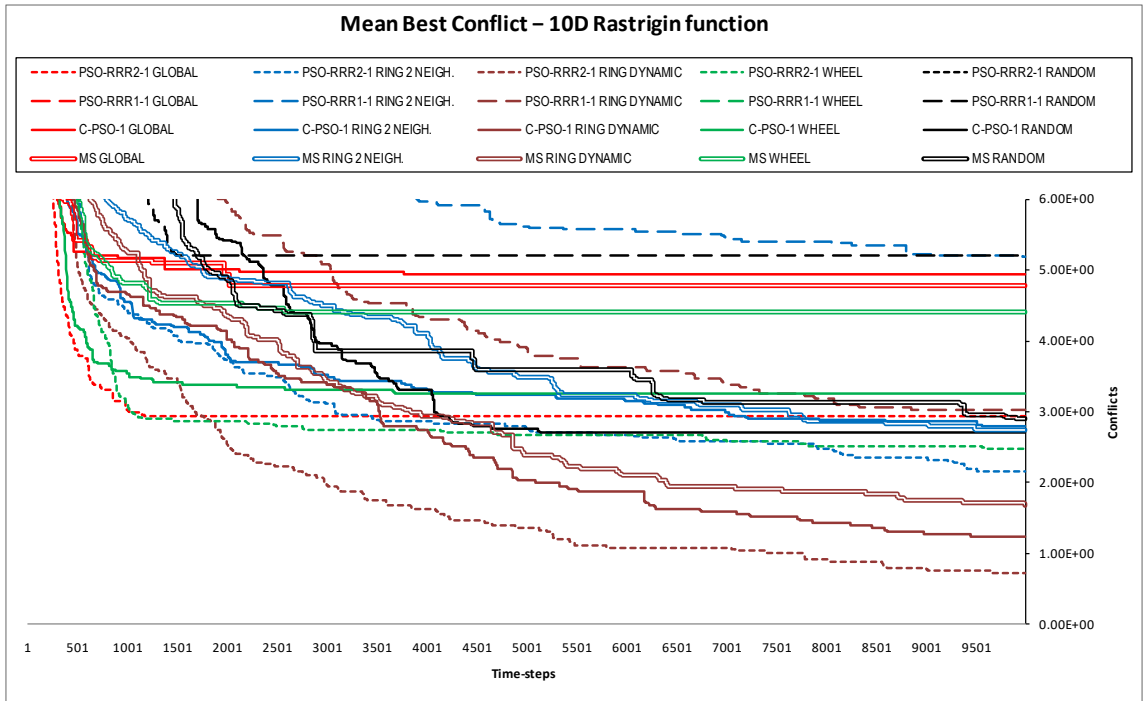

**Figure 14. Convergence curves of the mean best conflict for the 10D Rastrigin function, associated to Table 8. The colour-codes used to identify the neighbourhood structures are the same in the table and figure associated.**

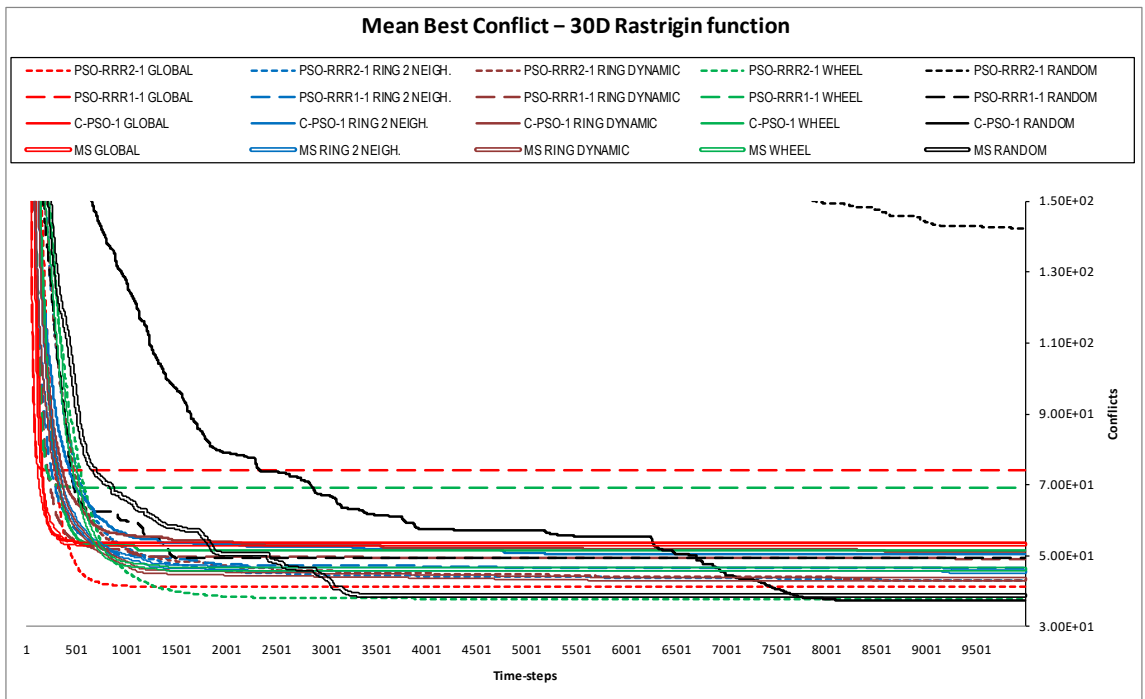

**Figure 15. Convergence curves of the mean best conflict for the 30D Rastrigin function, associated to Table 9. The colour-codes used to identify the neighbourhood structures are the same in the table and figure associated.**

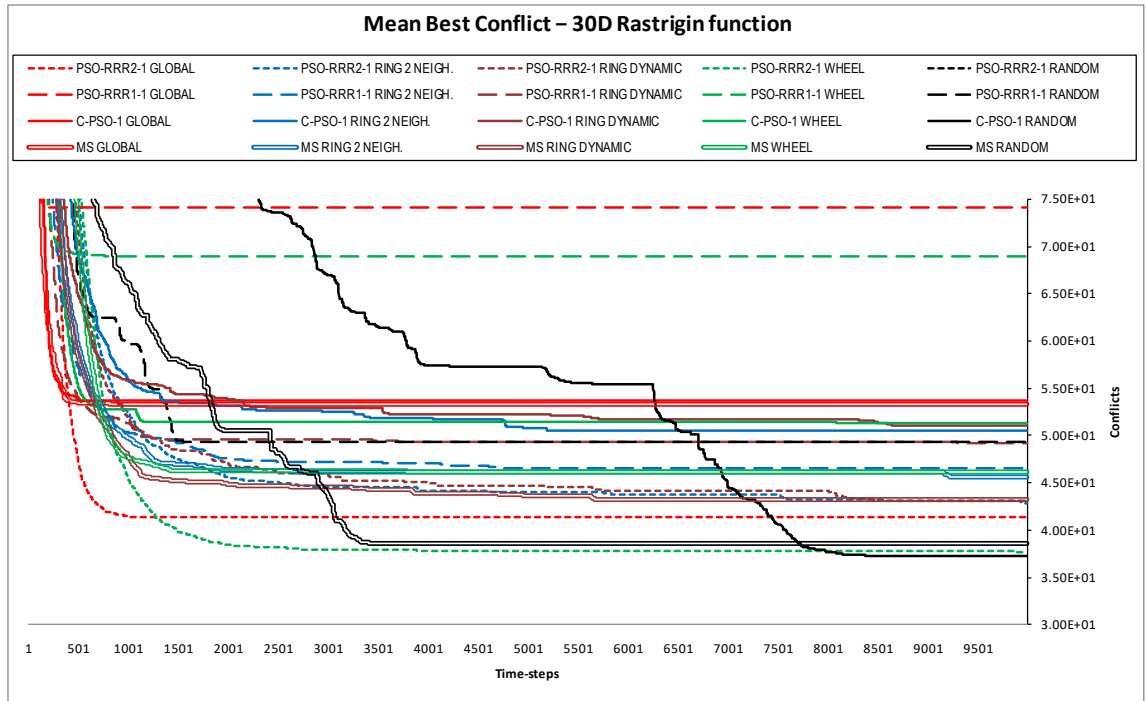

**Figure 16. Convergence curves of the mean best conflict for the 30D Rastrigin function, associated to Table 9. The colour-codes used to identify the neighbourhood structures are the same in the table and figure associated.**

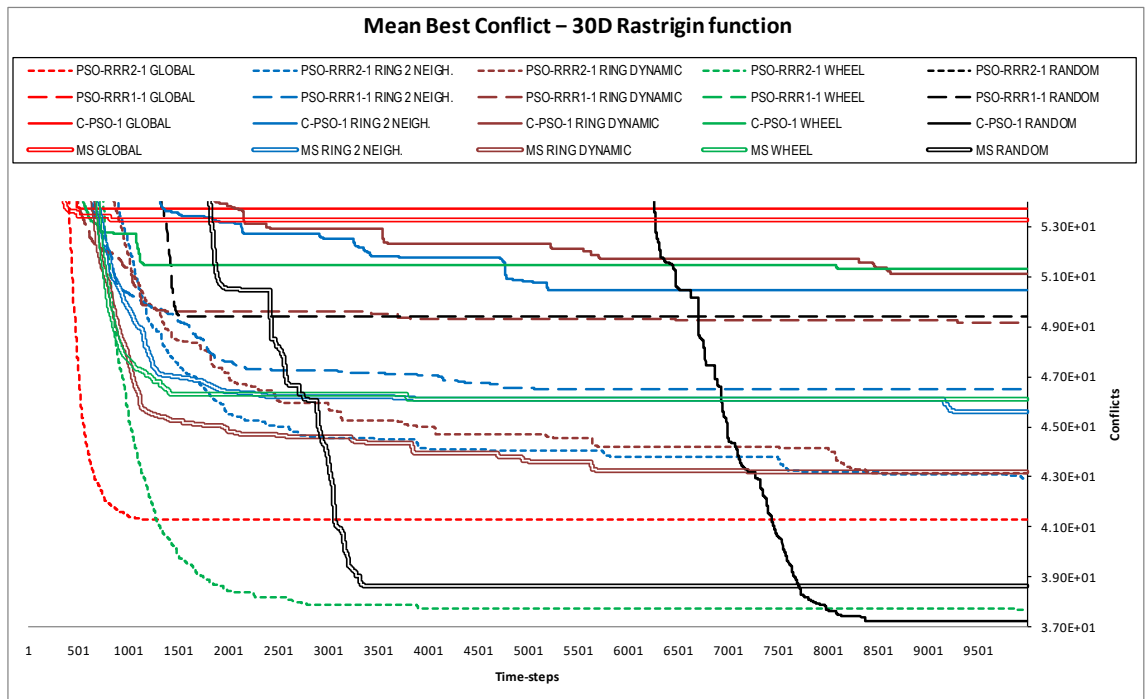

**Figure 17. Convergence curves of the mean best conflict for the 30D Rastrigin function, associated to Table 9. The colour-codes used to identify the neighbourhood structures are the same in the table and figure associated.**

**Table 10. Statistical results out of 25 runs for the PSO-RRR2-1, the PSO-RRR1-1, the C-PSO-1, and a Multi-Swarm algorithm optimizing the 2-dimensional Griewank function. The neighbourhoods tested are the GLOBAL; the RING with 2 neighbours; the RING with linearly increasing number of neighbours (from 2 to 'swarm-size – 1'); the WHEEL; and a RANDOM topology. A run with an error no greater than 0.0001 is regarded as successful.**

| OPTIMIZER | NEIGHBOURHOOD STRUCTURE | | Time-steps | GRIEWANK 2D | | | | OPTIMUM = 0 | |
|---|---|---|---|---|---|---|---|---|---|
| | | | | BEST | MEDIAN | MEAN | WORST | MEAN PB_ME | [%] Success |
| PSO-RRR2-1 | GLOBAL | | 10000 | 0.00E+00 | 0.00E+00 | 2.96E-04 | 7.40E-03 | 7.41E-12 | 96 |
| | | | 1000 | 0.00E+00 | 0.00E+00 | 5.92E-04 | 7.40E-03 | 1.10E-03 | - |
| | RING | nn = 2 | 10000 | 0.00E+00 | 0.00E+00 | 0.00E+00 | 0.00E+00 | 3.65E-04 | 100 |
| | | | 1000 | 0.00E+00 | 0.00E+00 | 0.00E+00 | 0.00E+00 | 2.10E-03 | - |
| | | nni = 2 nnf = (m – 1) | 10000 | 0.00E+00 | 0.00E+00 | 0.00E+00 | 0.00E+00 | 6.57E-12 | 100 |
| | | | 1000 | 0.00E+00 | 0.00E+00 | 0.00E+00 | 0.00E+00 | 1.64E-03 | - |
| | WHEEL | | 10000 | 0.00E+00 | 0.00E+00 | 0.00E+00 | 0.00E+00 | 3.57E-05 | 100 |
| | | | 1000 | 0.00E+00 | 0.00E+00 | 2.10E-04 | 4.65E-03 | 1.56E-03 | - |
| | RANDOM | | 10000 | 0.00E+00 | 0.00E+00 | 0.00E+00 | 0.00E+00 | 7.05E-122 | 100 |
| | | | 1000 | 0.00E+00 | 0.00E+00 | 1.70E-08 | 3.18E-07 | 1.84E-03 | - |
| PSO-RRR1-1 | GLOBAL | | 10000 | 0.00E+00 | 0.00E+00 | 5.92E-04 | 7.40E-03 | 5.76E-12 | 92 |
| | | | 1000 | 0.00E+00 | 0.00E+00 | 1.18E-03 | 9.08E-04 | 9.08E-04 | - |
| | RING | nn = 2 | 10000 | 0.00E+00 | 0.00E+00 | 0.00E+00 | 0.00E+00 | 4.15E-04 | 100 |
| | | | 1000 | 0.00E+00 | 0.00E+00 | 3.02E-13 | 7.54E-12 | 2.05E-03 | - |
| | | nni = 2 nnf = (m – 1) | 10000 | 0.00E+00 | 0.00E+00 | 0.00E+00 | 0.00E+00 | 6.45E-12 | 100 |
| | | | 1000 | 0.00E+00 | 0.00E+00 | 0.00E+00 | 0.00E+00 | 1.67E-03 | - |
| | WHEEL | | 10000 | 0.00E+00 | 0.00E+00 | 8.88E-04 | 7.40E-03 | 2.96E-05 | 88 |
| | | | 1000 | 0.00E+00 | 0.00E+00 | 9.21E-04 | 7.96E-03 | 9.56E-04 | - |
| | RANDOM | | 10000 | 0.00E+00 | 0.00E+00 | 0.00E+00 | 0.00E+00 | 7.32E-12 | 100 |
| | | | 1000 | 0.00E+00 | 0.00E+00 | 0.00E+00 | 0.00E+00 | 8.54E-04 | - |
| C-PSO-1 | GLOBAL | | 10000 | 0.00E+00 | 0.00E+00 | 0.00E+00 | 0.00E+00 | 6.40E-12 | 100 |
| | | | 1000 | 0.00E+00 | 0.00E+00 | 1.18E-03 | 7.40E-03 | 9.46E-04 | - |
| | RING | nn = 2 | 10000 | 0.00E+00 | 0.00E+00 | 0.00E+00 | 0.00E+00 | 1.57E-04 | 100 |
| | | | 1000 | 0.00E+00 | 0.00E+00 | 0.00E+00 | 0.00E+00 | 1.96E-03 | - |
| | | nni = 2 nnf = (m – 1) | 10000 | 0.00E+00 | 0.00E+00 | 0.00E+00 | 0.00E+00 | 6.81E-12 | 100 |
| | | | 1000 | 0.00E+00 | 0.00E+00 | 0.00E+00 | 0.00E+00 | 1.67E-03 | - |
| | WHEEL | | 10000 | 0.00E+00 | 0.00E+00 | 0.00E+00 | 0.00E+00 | 2.55E-09 | 100 |
| | | | 1000 | 0.00E+00 | 0.00E+00 | 3.99E-04 | 7.40E-03 | 1.37E-03 | - |
| | RANDOM | | 10000 | 0.00E+00 | 0.00E+00 | 0.00E+00 | 0.00E+00 | 6.50E-12 | 100 |
| | | | 1000 | 0.00E+00 | 0.00E+00 | 3.60E-12 | 9.00E-11 | 1.01E-03 | - |
| Multi-Swarm | GLOBAL | | 10000 | 0.00E+00 | 0.00E+00 | 0.00E+00 | 0.00E+00 | 6.56E-12 | 100 |
| | | | 1000 | 0.00E+00 | 0.00E+00 | 5.95E-16 | 1.49E-14 | 8.92E-04 | - |
| | RING | nn = 2 | 10000 | 0.00E+00 | 0.00E+00 | 0.00E+00 | 0.00E+00 | 2.51E-04 | 100 |
| | | | 1000 | 0.00E+00 | 0.00E+00 | 2.39E-05 | 5.88E-04 | 2.05E-03 | - |
| | | nni = 2 nnf = (m – 1) | 10000 | 0.00E+00 | 0.00E+00 | 0.00E+00 | 0.00E+00 | 6.74E-12 | 100 |
| | | | 1000 | 0.00E+00 | 0.00E+00 | 0.00E+00 | 0.00E+00 | 1.59E-03 | - |
| | WHEEL | | 10000 | 0.00E+00 | 0.00E+00 | 0.00E+00 | 0.00E+00 | 6.51E-12 | 100 |
| | | | 1000 | 0.00E+00 | 0.00E+00 | 2.99E-04 | 7.40E-03 | 1.25E-03 | - |
| | RANDOM | | 10000 | 0.00E+00 | 0.00E+00 | 0.00E+00 | 0.00E+00 | 6.64E-12 | 100 |
| | | | 1000 | 0.00E+00 | 0.00E+00 | 0.00E+00 | 0.00E+00 | 1.16E-03 | - |

**Table 11. Statistical results out of 25 runs for the PSO-RRR2-1, the PSO-RRR1-1, the C-PSO-1, and a Multi-Swarm algorithm optimizing the 10-dimensional Griewank function. The neighbourhoods tested are the GLOBAL; the RING with 2 neighbours; the RING with linearly increasing number of neighbours (from 2 to 'swarm-size – 1'); the WHEEL; and a RANDOM topology. A run with an error no greater than 0.0001 is regarded as successful.**

| OPTIMIZER | NEIGHBOURHOOD STRUCTURE | | Time-steps | GRIEWANK 10D | | | | OPTIMUM = 0 | |
|---|---|---|---|---|---|---|---|---|---|
| | | | | BEST | MEDIAN | MEAN | WORST | MEAN PB_ME | [%] Success |
| PSO-RRR2-1 | GLOBAL | | 10000 | 1.97E-02 | 5.66E-02 | 6.81E-02 | 1.43E-01 | 4.94E-07 | 0 |
| | | | 1000 | 1.97E-02 | 6.16E-02 | 7.14E-02 | 1.43E-01 | 1.18E-04 | - |
| | RING | nn = 2 | 10000 | 0.00E+00 | 2.46E-02 | 2.66E-02 | 6.15E-02 | 1.81E-03 | 4 |
| | | | 1000 | 2.96E-07 | 2.95E-02 | 3.41E-02 | 6.88E-02 | 2.01E-03 | - |
| | | nni = 2 nnf = (m – 1) | 10000 | 7.40E-03 | 2.96E-02 | 3.02E-02 | 5.65E-02 | 1.35E-03 | 0 |
| | | | 1000 | 9.86E-03 | 3.94E-02 | 3.68E-02 | 5.66E-02 | 1.54E-03 | - |
| | WHEEL | | 10000 | 7.40E-03 | 5.66E-02 | 5.98E-02 | 1.48E-01 | 6.64E-05 | 0 |
| | | | 1000 | 7.40E-03 | 6.64E-02 | 7.03E-02 | 1.52E-01 | 4.81E-04 | - |
| | RANDOM | | 10000 | 3.20E-02 | 4.48E-01 | 3.97E-01 | 7.91E-01 | 2.89E-03 | 0 |
| | | | 1000 | 1.23E-01 | 6.21E-01 | 6.02E-01 | 2.99E-03 | 2.99E-03 | - |
| PSO-RRR1-1 | GLOBAL | | 10000 | 2.96E-02 | 9.11E-02 | 9.27E-02 | 1.82E-01 | 1.81E-12 | 0 |
| | | | 1000 | 2.96E-02 | 9.11E-02 | 9.27E-02 | 1.82E-01 | 1.48E-05 | - |
| | RING | nn = 2 | 10000 | 0.00E+00 | 3.19E-02 | 3.05E-02 | 7.38E-02 | 1.58E-03 | 8 |
| | | | 1000 | 0.00E+00 | 3.94E-02 | 3.64E-02 | 7.62E-02 | 1.66E-03 | - |
| | | nni = 2 nnf = (m – 1) | 10000 | 0.00E+00 | 2.22E-02 | 2.95E-02 | 6.64E-02 | 1.27E-03 | 12 |
| | | | 1000 | 9.86E-03 | 4.18E-02 | 4.26E-02 | 1.11E-01 | 1.61E-03 | - |
| | WHEEL | | 10000 | 1.97E-02 | 9.11E-02 | 1.24E-01 | 3.84E-01 | 1.24E-12 | 0 |
| | | | 1000 | 1.97E-02 | 9.11E-02 | 1.27E-01 | 3.84E-01 | 6.60E-05 | - |
| | RANDOM | | 10000 | 1.72E-02 | 5.66E-02 | 5.60E-02 | 9.35E-02 | 1.07E-04 | 0 |
| | | | 1000 | 2.95E-02 | 6.89E-02 | 1.09E-01 | 5.28E-01 | 4.53E-04 | - |
| C-PSO-1 | GLOBAL | | 10000 | 1.97E-02 | 6.64E-02 | 6.68E-02 | 1.38E-01 | 1.65E-06 | 0 |
| | | | 1000 | 2.71E-02 | 6.89E-02 | 7.21E-02 | 1.38E-01 | 1.22E-04 | - |
| | RING | nn = 2 | 10000 | 0.00E+00 | 2.46E-02 | 2.36E-02 | 4.68E-02 | 1.55E-03 | 4 |
| | | | 1000 | 0.00E+00 | 2.71E-02 | 2.85E-02 | 8.87E-02 | 1.66E-03 | - |
| | | nni = 2 nnf = (m – 1) | 10000 | 0.00E+00 | 2.71E-02 | 2.49E-02 | 5.91E-02 | 1.19E-03 | 4 |
| | | | 1000 | 7.40E-03 | 3.69E-02 | 3.52E-02 | 7.38E-02 | 1.52E-03 | - |
| | WHEEL | | 10000 | 0.00E+00 | 6.65E-02 | 6.65E-02 | 1.30E-01 | 2.23E-05 | 4 |
| | | | 1000 | 1.97E-02 | 7.13E-02 | 7.34E-02 | 1.30E-01 | 4.00E-04 | - |
| | RANDOM | | 10000 | 9.86E-03 | 7.13E-02 | 1.58E-01 | 4.75E-01 | 6.70E-04 | 0 |
| | | | 1000 | 2.21E-02 | 4.25E-01 | 3.69E-01 | 7.91E-01 | 1.47E-03 | - |
| Multi-Swarm | GLOBAL | | 10000 | 1.48E-02 | 6.64E-02 | 6.64E-02 | 1.38E-01 | 1.48E-05 | 0 |
| | | | 1000 | 2.95E-02 | 7.13E-02 | 7.85E-02 | 1.85E-01 | 1.40E-04 | - |
| | RING | nn = 2 | 10000 | 0.00E+00 | 1.97E-02 | 2.15E-02 | 5.90E-02 | 1.60E-03 | 8 |
| | | | 1000 | 7.40E-03 | 2.22E-02 | 2.71E-02 | 5.90E-02 | 1.75E-03 | - |
| | | nni = 2 nnf = (m – 1) | 10000 | 0.00E+00 | 2.46E-02 | 2.77E-02 | 6.89E-02 | 1.47E-03 | 8 |
| | | | 1000 | 0.00E+00 | 3.45E-02 | 3.59E-02 | 7.38E-02 | 1.74E-03 | - |
| | WHEEL | | 10000 | 1.97E-02 | 6.15E-02 | 6.50E-02 | 1.45E-01 | 1.70E-05 | 0 |
| | | | 1000 | 4.55E-02 | 7.38E-02 | 8.35E-02 | 1.45E-01 | 3.40E-04 | - |
| | RANDOM | | 10000 | 1.72E-02 | 3.94E-02 | 5.85E-02 | 2.49E-01 | 7.33E-05 | 0 |
| | | | 1000 | 1.72E-02 | 6.90E-02 | 1.28E-01 | 6.89E-01 | 6.48E-04 | - |

**Table 12. Statistical results out of 25 runs for the PSO-RRR2-1, the PSO-RRR1-1, the C-PSO-1, and a Multi-Swarm algorithm optimizing the 30-dimensional Griewank function. The neighbourhoods tested are the GLOBAL; the RING with 2 neighbours; the RING with linearly increasing number of neighbours (from 2 to 'swarm-size – 1'); the WHEEL; and a RANDOM topology. A run with an error no greater than 0.0001 is regarded as successful.**

| OPTIMIZER | NEIGHBOURHOOD STRUCTURE | | Time-steps | GRIEWANK 30D | | | | OPTIMUM = 0 | |
|---|---|---|---|---|---|---|---|---|---|
| | | | | BEST | MEDIAN | MEAN | WORST | MEAN PB_ME | [%] Success |
| PSO-RRR2-1 | GLOBAL | | 10000 | 0.00E+00 | 7.40E-03 | 9.35E-03 | 2.96E-02 | 4.23E-12 | 44 |
| | | | 1000 | 7.95E-06 | 7.44E-03 | 9.40E-02 | 2.96E-02 | 2.93E-06 | - |
| | RING | nn = 2 | 10000 | 0.00E+00 | 0.00E+00 | 2.96E-04 | 7.40E-03 | 1.22E-06 | 96 |
| | | | 1000 | 2.54E-01 | 4.15E-01 | 4.23E-01 | 6.79E-01 | 2.52E-04 | - |
| | | nni = 2 nnf = (m – 1) | 10000 | 0.00E+00 | 0.00E+00 | 4.05E-03 | 1.72E-02 | 3.77E-06 | 64 |
| | | | 1000 | 3.65E-02 | 7.18E-02 | 8.05E-02 | 1.70E-01 | 1.13E-04 | - |
| | WHEEL | | 10000 | 0.00E+00 | 7.40E-03 | 1.03E-02 | 3.94E-02 | 2.15E-12 | 40 |
| | | | 1000 | 6.00E-02 | 2.26E-01 | 2.86E-01 | 7.71E-01 | 9.81E-05 | - |
| | RANDOM | | 10000 | 6.05E-04 | 3.52E-02 | 1.63E-01 | 1.02E+00 | 1.38E-04 | 0 |
| | | | 1000 | 3.47E+00 | 1.02E+01 | 1.23E+01 | 3.14E+01 | 1.01E-02 | - |
| PSO-RRR1-1 | GLOBAL | | 10000 | 3.29E-08 | 6.46E-02 | 1.02E-01 | 7.40E-01 | 3.91E-13 | 4 |
| | | | 1000 | 3.29E-08 | 7.11E-02 | 1.12E-01 | 7.40E-01 | 4.04E-14 | - |
| | RING | nn = 2 | 10000 | 0.00E+00 | 0.00E+00 | 6.90E-04 | 9.86E-03 | 6.40E-08 | 92 |
| | | | 1000 | 4.02E-11 | 7.73E-10 | 6.91E-04 | 9.86E-03 | 1.38E-06 | - |
| | | nni = 2 nnf = (m – 1) | 10000 | 0.00E+00 | 0.00E+00 | 1.58E-03 | 9.86E-03 | 8.49E-09 | 80 |
| | | | 1000 | 0.00E+00 | 4.76E-08 | 3.15E-03 | 1.48E-02 | 1.11E-06 | - |
| | WHEEL | | 10000 | 4.58E-13 | 8.57E-02 | 1.46E-01 | 8.89E-01 | 2.96E-13 | 4 |
| | | | 1000 | 7.63E-03 | 9.45E-02 | 1.77E-01 | 9.38E-01 | 1.72E-06 | - |
| | RANDOM | | 10000 | 0.00E+00 | 9.86E-03 | 1.47E-02 | 7.36E-02 | 4.13E-12 | 28 |
| | | | 1000 | 0.00E+00 | 9.86E-03 | 1.47E-02 | 7.36E-02 | 3.71E-12 | - |
| C-PSO-1 | GLOBAL | | 10000 | 0.00E+00 | 1.23E-02 | 1.79E-02 | 7.09E-02 | 2.56E-12 | 36 |
| | | | 1000 | 0.00E+00 | 1.23E-02 | 1.79E-02 | 7.09E-02 | 2.93E-12 | - |
| | RING | nn = 2 | 10000 | 0.00E+00 | 0.00E+00 | 0.00E+00 | 0.00E+00 | 1.14E-07 | 100 |
| | | | 1000 | 4.55E-06 | 3.98E-05 | 2.60E-03 | 2.22E-02 | 1.52E-05 | - |
| | | nni = 2 nnf = (m – 1) | 10000 | 0.00E+00 | 0.00E+00 | 1.97E-03 | 1.72E-02 | 4.73E-07 | 84 |
| | | | 1000 | 3.22E-09 | 3.49E-08 | 2.37E-03 | 1.72E-02 | 8.37E-06 | - |
| | WHEEL | | 10000 | 0.00E+00 | 9.86E-03 | 3.55E-02 | 2.37E-01 | 1.68E-12 | 24 |
| | | | 1000 | 7.76E-08 | 9.87E-03 | 3.55E-02 | 2.37E-01 | 1.62E-07 | - |
| | RANDOM | | 10000 | 0.00E+00 | 7.40E-03 | 9.15E-03 | 5.15E-02 | 5.03E-12 | 44 |
| | | | 1000 | 3.95E-08 | 7.40E-03 | 9.15E-03 | 5.16E-02 | 4.29E-07 | - |
| Multi-Swarm | GLOBAL | | 10000 | 0.00E+00 | 4.67E-02 | 5.18E-02 | 1.41E-01 | 2.42E-12 | 4 |
| | | | 1000 | 6.66E-16 | 4.67E-02 | 5.18E-02 | 1.41E-01 | 4.06E-08 | - |
| | RING | nn = 2 | 10000 | 0.00E+00 | 0.00E+00 | 2.17E-03 | 1.23E-02 | 1.10E-06 | 76 |
| | | | 1000 | 6.02E-08 | 7.40E-03 | 6.97E-03 | 3.92E-02 | 3.08E-05 | - |
| | | nni = 2 nnf = (m – 1) | 10000 | 0.00E+00 | 0.00E+00 | 6.39E-03 | 3.92E-02 | 4.12E-12 | 64 |
| | | | 1000 | 9.47E-11 | 7.40E-03 | 9.54E-03 | 3.92E-02 | 1.26E-05 | - |
| | WHEEL | | 10000 | 0.00E+00 | 1.23E-02 | 1.75E-02 | 1.30E-01 | 1.66E-12 | 44 |
| | | | 1000 | 4.12E-03 | 5.41E-02 | 1.10E-01 | 4.16E-01 | 3.36E-05 | - |
| | RANDOM | | 10000 | 0.00E+00 | 7.40E-03 | 1.43E-02 | 7.09E-02 | 5.23E-12 | 44 |
| | | | 1000 | 2.66E-15 | 7.40E-03 | 1.43E-02 | 7.09E-02 | 2.14E-07 | - |

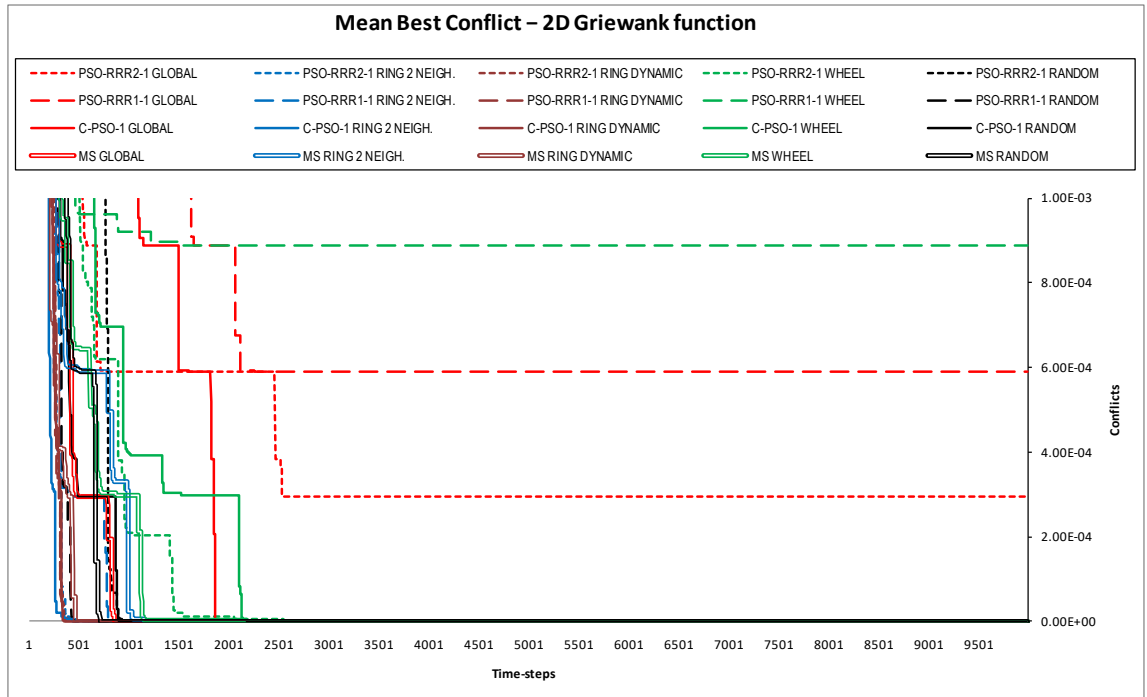

**Figure 18. Convergence curves of the mean best conflict for the 2D Griewank function, associated to Table 10. The colour-codes used to identify the neighbourhood structures are the same in the table and figure associated.**

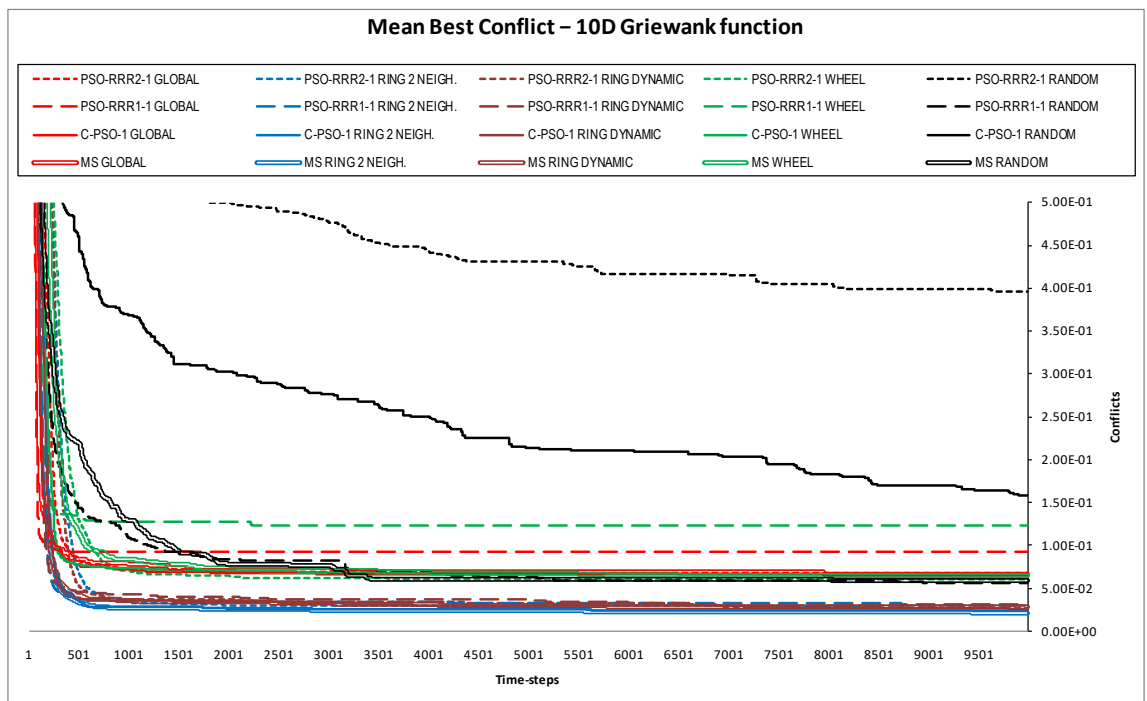

**Figure 19. Convergence curves of the mean best conflict for the 10D Griewank function, associated to Table 11. The colour-codes used to identify the neighbourhood structures are the same in the table and figure associated.**

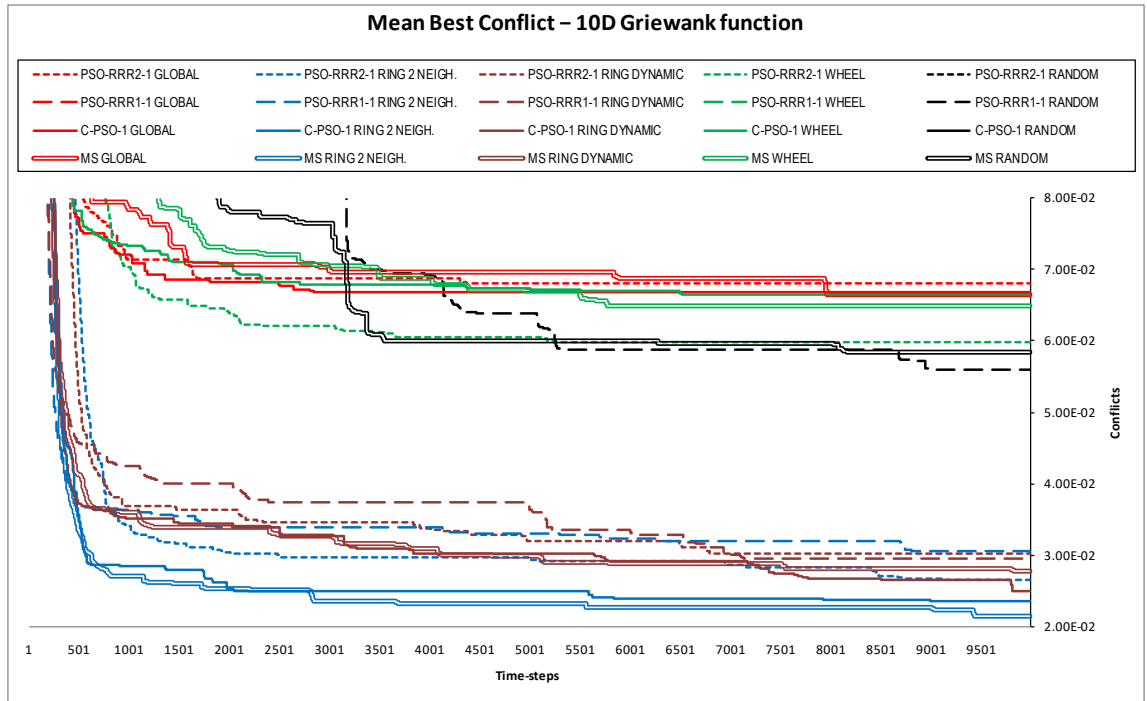

**Figure 20. Convergence curves of the mean best conflict for the 10D Griewank function, associated to Table 11. The colour-codes used to identify the neighbourhood structures are the same in the table and figure associated.**

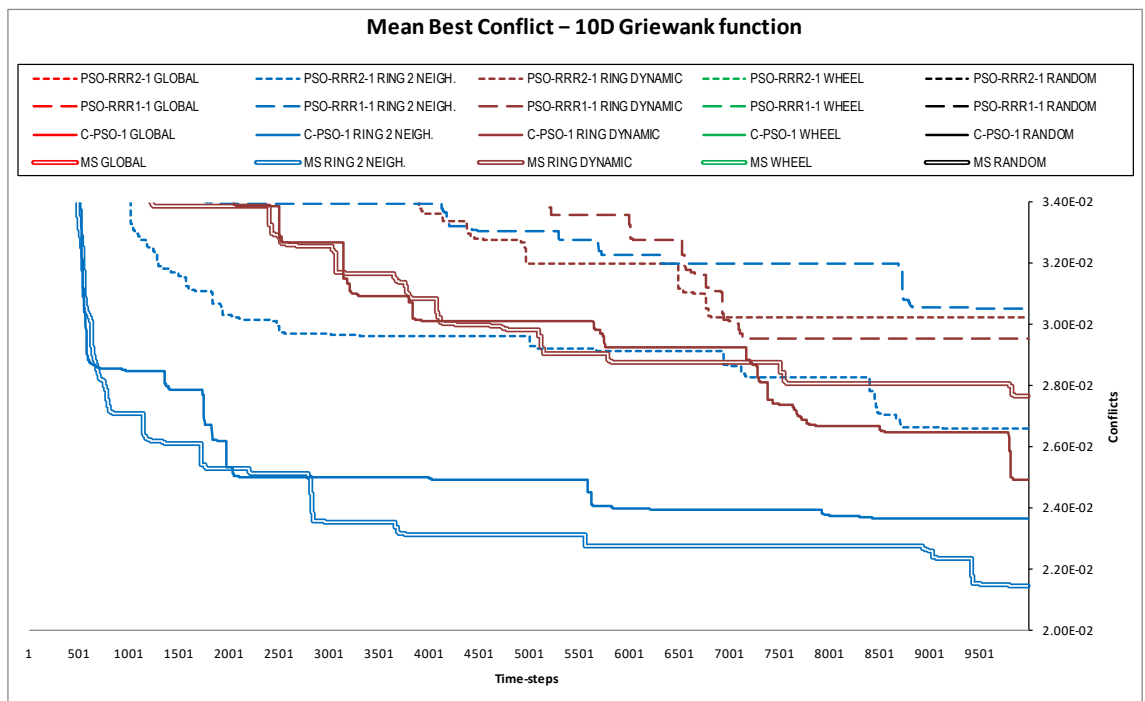

**Figure 21. Convergence curves of the mean best conflict for the 10D Griewank function, associated to Table 11. The colour-codes used to identify the neighbourhood structures are the same in the table and figure associated.**

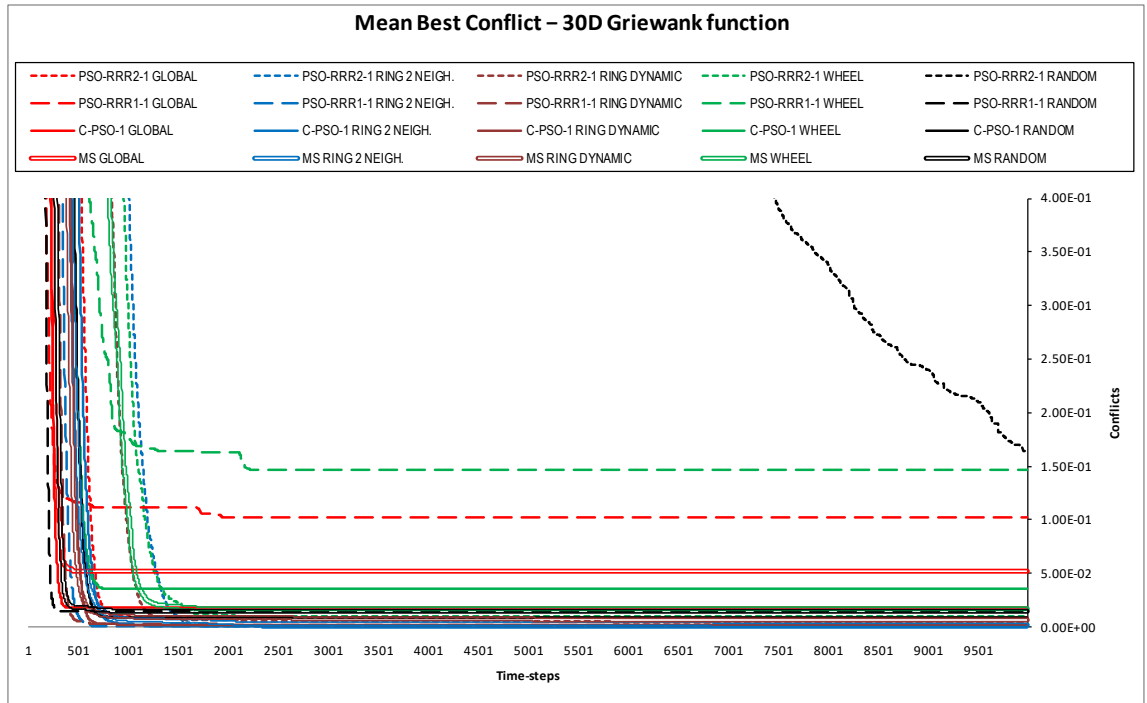

**Figure 22. Convergence curves of the mean best conflict for the 30D Griewank function, associated to Table 12. The colour-codes used to identify the neighbourhood structures are the same in the table and figure associated.**

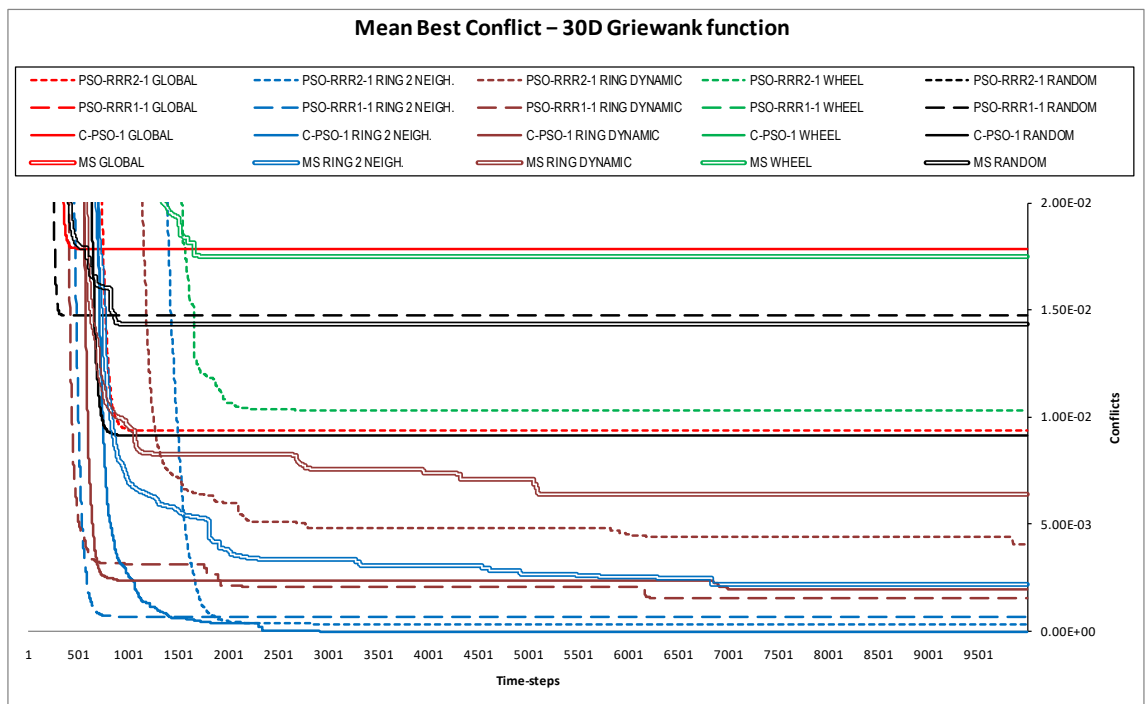

**Figure 23. Convergence curves of the mean best conflict for the 30D Griewank function, associated to Table 12. The colour-codes used to identify the neighbourhood structures are the same in the table and figure associated.**

**Table 13. Statistical results out of 25 runs for the PSO-RRR2-1, the PSO-RRR1-1, the C-PSO-1, and a Multi-Swarm algorithm optimizing the 2-dimensional Schaffer f6 function. The neighbourhoods tested are the GLOBAL; the RING with 2 neighbours; the RING with linearly increasing number of neighbours (from 2 to 'swarm-size – 1'); the WHEEL; and a RANDOM topology. A run with an error no greater than 0.0001 is regarded as successful.**

| OPTIMIZER | NEIGHBOURHOOD STRUCTURE | | Time-steps | SCHAFFER F6 2D | | | | OPTIMUM = 0 | |
|---|---|---|---|---|---|---|---|---|---|
| | | | | BEST | MEDIAN | MEAN | WORST | MEAN PB_ME | [%] Success |
| PSO-RRR2-1 | GLOBAL | | 10000 | 0.00E+00 | 0.00E+00 | 3.89E-04 | 9.72E-03 | 2.58E-05 | 96 |
| | | | 1000 | 0.00E+00 | 0.00E+00 | 7.77E-04 | 9.72E-03 | 2.79E-03 | - |
| | RING | nn = 2 | 10000 | 0.00E+00 | 0.00E+00 | 0.00E+00 | 0.00E+00 | 3.60E-04 | 100 |
| | | | 1000 | 0.00E+00 | 0.00E+00 | 3.89E-04 | 9.72E-03 | 6.82E-03 | - |
| | | nni = 2 nnf = (m – 1) | 10000 | 0.00E+00 | 0.00E+00 | 0.00E+00 | 0.00E+00 | 1.25E-11 | 100 |
| | | | 1000 | 0.00E+00 | 0.00E+00 | 5.43E-04 | 9.72E-03 | 5.57E-03 | - |
| | WHEEL | | 10000 | 0.00E+00 | 0.00E+00 | 7.77E-04 | 9.72E-03 | 7.57E-04 | 92 |
| | | | 1000 | 0.00E+00 | 0.00E+00 | 3.11E-03 | 9.72E-03 | 5.30E-03 | - |
| | RANDOM | | 10000 | 0.00E+00 | 0.00E+00 | 0.00E+00 | 0.00E+00 | 1.18E-11 | 100 |
| | | | 1000 | 0.00E+00 | 0.00E+00 | 2.75E-16 | 5.88E-15 | 5.00E-03 | - |
| PSO-RRR1-1 | GLOBAL | | 10000 | 0.00E+00 | 0.00E+00 | 1.17E-03 | 9.72E-03 | 1.13E-04 | 88 |
| | | | 1000 | 0.00E+00 | 0.00E+00 | 1.17E-03 | 9.72E-03 | 1.61E-03 | - |
| | RING | nn = 2 | 10000 | 0.00E+00 | 0.00E+00 | 0.00E+00 | 0.00E+00 | 1.29E-03 | 100 |
| | | | 1000 | 0.00E+00 | 0.00E+00 | 1.96E-03 | 9.72E-03 | 7.78E-03 | - |
| | | nni = 2 nnf = (m – 1) | 10000 | 0.00E+00 | 0.00E+00 | 0.00E+00 | 0.00E+00 | 1.26E-11 | 100 |
| | | | 1000 | 0.00E+00 | 0.00E+00 | 7.03E-07 | 1.76E-05 | 5.62E-03 | - |
| | WHEEL | | 10000 | 0.00E+00 | 0.00E+00 | 2.72E-03 | 9.72E-03 | 9.53E-04 | 72 |
| | | | 1000 | 0.00E+00 | 0.00E+00 | 3.50E-03 | 9.72E-03 | 3.96E-03 | - |
| | RANDOM | | 10000 | 0.00E+00 | 0.00E+00 | 0.00E+00 | 0.00E+00 | 1.14E-11 | 100 |
| | | | 1000 | 0.00E+00 | 0.00E+00 | 0.00E+00 | 0.00E+00 | 2.79E-03 | - |
| C-PSO-1 | GLOBAL | | 10000 | 0.00E+00 | 0.00E+00 | 1.17E-03 | 9.72E-03 | 2.01E-04 | 88 |
| | | | 1000 | 0.00E+00 | 0.00E+00 | 1.95E-03 | 9.72E-03 | 2.42E-03 | - |
| | RING | nn = 2 | 10000 | 0.00E+00 | 0.00E+00 | 3.89E-04 | 9.72E-03 | 1.38E-03 | 96 |
| | | | 1000 | 0.00E+00 | 0.00E+00 | 2.07E-03 | 9.72E-03 | 7.76E-03 | - |
| | | nni = 2 nnf = (m – 1) | 10000 | 0.00E+00 | 0.00E+00 | 0.00E+00 | 0.00E+00 | 1.23E-11 | 100 |
| | | | 1000 | 0.00E+00 | 0.00E+00 | 7.58E-06 | 1.21E-04 | 6.07E-03 | - |
| | WHEEL | | 10000 | 0.00E+00 | 0.00E+00 | 7.77E-04 | 9.72E-03 | 3.75E-04 | 92 |
| | | | 1000 | 0.00E+00 | 0.00E+00 | 2.76E-03 | 9.72E-03 | 4.62E-03 | - |
| | RANDOM | | 10000 | 0.00E+00 | 0.00E+00 | 0.00E+00 | 0.00E+00 | 1.19E-11 | 100 |
| | | | 1000 | 0.00E+00 | 0.00E+00 | 6.26E-04 | 9.72E-03 | 3.26E-03 | - |
| Multi-Swarm | GLOBAL | | 10000 | 0.00E+00 | 0.00E+00 | 2.33E-03 | 9.72E-03 | 1.05E-04 | 76 |
| | | | 1000 | 0.00E+00 | 0.00E+00 | 3.11E-03 | 9.72E-03 | 2.58E-03 | - |
| | RING | nn = 2 | 10000 | 0.00E+00 | 0.00E+00 | 0.00E+00 | 0.00E+00 | 7.13E-04 | 100 |
| | | | 1000 | 0.00E+00 | 0.00E+00 | 7.84E-04 | 9.72E-03 | 7.34E-03 | - |
| | | nni = 2 nnf = (m – 1) | 10000 | 0.00E+00 | 0.00E+00 | 0.00E+00 | 0.00E+00 | 1.36E-11 | 100 |
| | | | 1000 | 0.00E+00 | 0.00E+00 | 3.89E-04 | 9.72E-03 | 6.25E-03 | - |
| | WHEEL | | 10000 | 0.00E+00 | 0.00E+00 | 7.77E-04 | 9.72E-03 | 3.08E-04 | 92 |
| | | | 1000 | 0.00E+00 | 0.00E+00 | 1.24E-03 | 9.72E-03 | 4.38E-03 | - |
| | RANDOM | | 10000 | 0.00E+00 | 0.00E+00 | 0.00E+00 | 0.00E+00 | 1.13E-11 | 100 |
| | | | 1000 | 0.00E+00 | 0.00E+00 | 0.00E+00 | 0.00E+00 | 3.56E-03 | - |

**Table 14. Statistical results out of 25 runs for the PSO-RRR2-1, the PSO-RRR1-1, the C-PSO-1, and a Multi-Swarm algorithm optimizing the 10-dimensional Schaffer f6 function. The neighbourhoods tested are the GLOBAL; the RING with 2 neighbours; the RING with linearly increasing number of neighbours (from 2 to 'swarm-size – 1'); the WHEEL; and a RANDOM topology. A run with an error no greater than 0.0001 is regarded as successful.**

| OPTIMIZER | NEIGHBOURHOOD STRUCTURE | | Time-steps | SCHAFFER F6 10D | | | | OPTIMUM = 0 | |
|---|---|---|---|---|---|---|---|---|---|
| | | | | BEST | MEDIAN | MEAN | WORST | MEAN PB_ME | [%] Success |
| PSO-RRR2-1 | GLOBAL | | 10000 | 9.72E-03 | 9.72E-03 | 1.85E-02 | 3.72E-02 | 5.10E-04 | 0 |
| | | | 1000 | 9.72E-03 | 9.72E-03 | 2.18E-02 | 3.72E-02 | 1.21E-03 | - |
| | RING | nn = 2 | 10000 | 9.72E-03 | 9.72E-03 | 9.72E-03 | 9.72E-03 | 1.94E-03 | 0 |
| | | | 1000 | 9.72E-03 | 9.72E-03 | 1.08E-02 | 3.72E-02 | 3.24E-03 | - |
| | | nni = 2 nnf = (m – 1) | 10000 | 9.72E-03 | 9.72E-03 | 9.72E-03 | 9.72E-03 | 1.65E-03 | 0 |
| | | | 1000 | 9.72E-03 | 9.72E-03 | 1.19E-02 | 3.72E-02 | 2.13E-03 | - |
| | WHEEL | | 10000 | 9.72E-03 | 9.72E-03 | 1.52E-02 | 3.72E-02 | 1.40E-03 | 0 |
| | | | 1000 | 9.72E-03 | 9.72E-03 | 1.85E-02 | 3.72E-02 | 1.94E-03 | - |
| | RANDOM | | 10000 | 9.72E-03 | 9.72E-03 | 1.19E-02 | 3.72E-02 | 2.57E-03 | 0 |
| | | | 1000 | 3.72E-02 | 7.82E-02 | 7.24E-02 | 2.28E-01 | 8.20E-03 | - |
| PSO-RRR1-1 | GLOBAL | | 10000 | 9.72E-03 | 3.72E-02 | 3.45E-02 | 7.82E-02 | 1.29E-04 | 0 |
| | | | 1000 | 9.72E-03 | 3.72E-02 | 3.45E-02 | 7.82E-02 | 1.29E-04 | - |
| | RING | nn = 2 | 10000 | 9.72E-03 | 9.72E-03 | 2.18E-02 | 3.72E-02 | 2.43E-03 | 0 |
| | | | 1000 | 9.72E-03 | 3.72E-02 | 3.17E-02 | 3.72E-02 | 3.38E-03 | - |
| | | nni = 2 nnf = (m – 1) | 10000 | 9.72E-03 | 9.72E-03 | 9.72E-03 | 9.72E-03 | 1.39E-03 | 0 |
| | | | 1000 | 9.72E-03 | 9.72E-03 | 1.74E-02 | 3.72E-02 | 2.35E-03 | - |
| | WHEEL | | 10000 | 3.72E-02 | 1.27E-01 | 1.22E-01 | 2.73E-01 | 1.04E-03 | 0 |
| | | | 1000 | 3.72E-02 | 1.27E-01 | 1.26E-01 | 2.73E-01 | 1.23E-03 | - |
| | RANDOM | | 10000 | 9.72E-03 | 9.72E-03 | 9.72E-03 | 9.72E-03 | 1.29E-03 | 0 |
| | | | 1000 | 9.72E-03 | 9.72E-03 | 9.72E-03 | 9.72E-03 | 1.44E-03 | - |
| C-PSO-1 | GLOBAL | | 10000 | 9.72E-03 | 9.72E-03 | 1.96E-02 | 3.72E-02 | 3.01E-04 | 0 |
| | | | 1000 | 9.72E-03 | 9.72E-03 | 2.18E-02 | 3.72E-02 | 7.06E-04 | - |
| | RING | nn = 2 | 10000 | 9.72E-03 | 9.72E-03 | 1.08E-02 | 3.72E-02 | 2.21E-03 | 0 |
| | | | 1000 | 9.72E-03 | 9.72E-03 | 1.85E-02 | 3.72E-02 | 3.19E-03 | - |
| | | nni = 2 nnf = (m – 1) | 10000 | 9.72E-03 | 9.72E-03 | 9.72E-03 | 9.72E-03 | 1.92E-03 | 0 |
| | | | 1000 | 9.72E-03 | 9.72E-03 | 1.08E-02 | 3.72E-02 | 2.11E-03 | - |
| | WHEEL | | 10000 | 9.72E-03 | 3.72E-02 | 2.79E-02 | 7.82E-02 | 1.09E-03 | 0 |
| | | | 1000 | 9.72E-03 | 3.72E-02 | 3.01E-02 | 7.82E-02 | 1.36E-03 | - |
| | RANDOM | | 10000 | 9.72E-03 | 9.72E-03 | 9.72E-03 | 9.72E-03 | 1.68E-03 | 0 |
| | | | 1000 | 9.72E-03 | 9.72E-03 | 9.72E-03 | 9.72E-03 | 1.90E-03 | - |
| Multi-Swarm | GLOBAL | | 10000 | 9.72E-03 | 3.72E-02 | 2.95E-02 | 3.72E-02 | 3.18E-04 | 0 |
| | | | 1000 | 9.72E-03 | 3.72E-02 | 3.06E-02 | 3.72E-02 | 6.67E-04 | - |
| | RING | nn = 2 | 10000 | 9.72E-03 | 9.72E-03 | 1.19E-02 | 3.72E-02 | 2.03E-03 | 0 |
| | | | 1000 | 9.72E-03 | 9.72E-03 | 1.74E-02 | 3.72E-02 | 3.03E-03 | - |
| | | nni = 2 nnf = (m – 1) | 10000 | 9.72E-03 | 9.72E-03 | 9.72E-03 | 9.72E-03 | 1.76E-03 | 0 |
| | | | 1000 | 9.72E-03 | 9.72E-03 | 1.08E-02 | 3.72E-02 | 1.95E-03 | - |
| | WHEEL | | 10000 | 9.72E-03 | 9.72E-03 | 1.96E-02 | 3.72E-02 | 8.52E-04 | 0 |
| | | | 1000 | 9.72E-03 | 9.72E-03 | 2.18E-02 | 3.72E-02 | 1.17E-03 | - |
| | RANDOM | | 10000 | 9.72E-03 | 9.72E-03 | 9.72E-03 | 9.72E-03 | 1.63E-03 | 0 |
| | | | 1000 | 9.72E-03 | 9.72E-03 | 9.72E-03 | 9.72E-03 | 1.81E-03 | - |

**Table 15. Statistical results out of 25 runs for the PSO-RRR2-1, the PSO-RRR1-1, the C-PSO-1, and a Multi-Swarm algorithm optimizing the 30-dimensional Schaffer f6 function. The neighbourhoods tested are the GLOBAL; the RING with 2 neighbours; the RING with linearly increasing number of neighbours (from 2 to 'swarm-size – 1'); the WHEEL; and a RANDOM topology. A run with an error no greater than 0.0001 is regarded as successful.**

| OPTIMIZER | NEIGHBOURHOOD STRUCTURE | | Time-steps | SCHAFFER F6 30D | | | | OPTIMUM = 0 | |
|---|---|---|---|---|---|---|---|---|---|
| | | | | BEST | MEDIAN | MEAN | WORST | MEAN PB_ME | [%] Success |
| PSO-RRR2-1 | GLOBAL | | 10000 | 3.72E-02 | 7.82E-02 | 9.22E-02 | 1.27E-01 | 3.12E-04 | 0 |
| | | | 1000 | 7.82E-02 | 1.27E-01 | 1.08E-01 | 1.78E-01 | 8.50E-04 | - |
| | RING | nn = 2 | 10000 | 3.72E-02 | 7.82E-02 | 6.18E-02 | 7.82E-02 | 1.77E-03 | 0 |
| | | | 1000 | 1.27E-01 | 1.96E-01 | 2.01E-01 | 2.29E-01 | 4.40E-03 | - |
| | | nni = 2 nnf = (m – 1) | 10000 | 3.72E-02 | 3.72E-02 | 3.72E-02 | 3.72E-02 | 7.48E-04 | 0 |
| | | | 1000 | 7.82E-02 | 1.27E-01 | 1.26E-01 | 2.04E-01 | 2.76E-03 | - |
| | WHEEL | | 10000 | 7.82E-02 | 1.27E-01 | 1.24E-01 | 2.28E-01 | 7.45E-04 | 0 |
| | | | 1000 | 1.27E-01 | 1.78E-01 | 1.90E-01 | 3.46E-01 | 1.40E-03 | - |
| | RANDOM | | 10000 | 2.73E-01 | 3.96E-01 | 3.74E-01 | 4.42E-01 | 8.30E-03 | 0 |
| | | | 1000 | 4.72E-01 | 4.90E-01 | 4.88E-01 | 4.97E-01 | 1.95E-02 | - |
| PSO-RRR1-1 | GLOBAL | | 10000 | 3.12E-01 | 4.30E-01 | 4.25E-01 | 4.85E-01 | 4.76E-05 | 0 |
| | | | 1000 | 3.12E-01 | 4.30E-01 | 4.26E-01 | 4.87E-01 | 2.01E-04 | - |
| | RING | nn = 2 | 10000 | 7.82E-02 | 1.78E-01 | 1.67E-01 | 2.73E-01 | 1.72E-03 | 0 |
| | | | 1000 | 1.27E-01 | 2.28E-01 | 2.14E-01 | 3.12E-01 | 2.38E-03 | - |
| | | nni = 2 nnf = (m – 1) | 10000 | 3.72E-02 | 1.27E-01 | 1.12E-01 | 2.28E-01 | 6.68E-04 | 0 |
| | | | 1000 | 7.82E-02 | 1.27E-01 | 1.49E-01 | 2.73E-01 | 1.23E-03 | - |
| | WHEEL | | 10000 | 4.52E-01 | 4.89E-01 | 4.87E-01 | 4.96E-01 | 4.22E-04 | 0 |
| | | | 1000 | 4.52E-01 | 4.89E-01 | 4.87E-01 | 4.96E-01 | 3.34E-04 | - |
| | RANDOM | | 10000 | 3.72E-02 | 3.72E-02 | 5.63E-02 | 1.78E-01 | 6.69E-04 | 0 |
| | | | 1000 | 3.72E-02 | 7.82E-02 | 8.58E-02 | 1.78E-01 | 9.84E-04 | - |
| C-PSO-1 | GLOBAL | | 10000 | 7.82E-02 | 1.27E-01 | 1.31E-01 | 2.73E-01 | 1.67E-04 | 0 |
| | | | 1000 | 7.82E-02 | 1.27E-01 | 1.40E-01 | 2.73E-01 | 4.94E-04 | - |
| | RING | nn = 2 | 10000 | 3.72E-02 | 3.72E-02 | 5.52E-02 | 7.82E-02 | 1.59E-03 | 0 |
| | | | 1000 | 1.27E-01 | 1.27E-01 | 1.52E-01 | 1.78E-01 | 3.23E-03 | - |
| | | nni = 2 nnf = (m – 1) | 10000 | 3.72E-02 | 3.72E-02 | 3.72E-02 | 3.72E-02 | 6.49E-04 | 0 |
| | | | 1000 | 7.82E-02 | 7.82E-02 | 9.19E-02 | 1.27E-01 | 2.00E-03 | - |
| | WHEEL | | 10000 | 1.27E-01 | 2.73E-01 | 2.63E-01 | 4.30E-01 | 8.05E-04 | 0 |
| | | | 1000 | 1.78E-01 | 2.73E-01 | 2.76E-01 | 4.30E-01 | 9.49E-04 | - |
| | RANDOM | | 10000 | 3.72E-02 | 3.72E-02 | 5.36E-02 | 7.82E-02 | 1.32E-03 | 0 |
| | | | 1000 | 1.78E-01 | 2.29E-01 | 2.54E-01 | 3.73E-01 | 4.54E-03 | - |
| Multi-Swarm | GLOBAL | | 10000 | 7.82E-02 | 1.78E-01 | 1.86E-01 | 2.73E-01 | 2.30E-04 | 0 |
| | | | 1000 | 1.27E-01 | 1.78E-01 | 1.93E-01 | 2.73E-01 | 5.26E-04 | - |
| | RING | nn = 2 | 10000 | 3.72E-02 | 7.82E-02 | 7.45E-02 | 1.27E-01 | 1.58E-03 | 0 |
| | | | 1000 | 1.27E-01 | 1.78E-01 | 1.68E-01 | 2.28E-01 | 3.42E-03 | - |
| | | nni = 2 nnf = (m – 1) | 10000 | 3.72E-02 | 3.72E-02 | 4.38E-02 | 7.82E-02 | 5.48E-04 | 0 |
| | | | 1000 | 3.74E-02 | 1.27E-01 | 1.08E-01 | 1.78E-01 | 1.89E-03 | - |
| | WHEEL | | 10000 | 7.82E-02 | 2.28E-01 | 2.13E-01 | 3.46E-01 | 6.95E-04 | 0 |
| | | | 1000 | 7.82E-02 | 2.73E-01 | 2.46E-01 | 3.46E-01 | 1.07E-03 | - |
| | RANDOM | | 10000 | 3.72E-02 | 3.72E-02 | 4.05E-02 | 7.82E-02 | 1.12E-03 | 0 |
| | | | 1000 | 7.82E-02 | 1.27E-01 | 1.10E-01 | 1.78E-01 | 2.18E-03 | - |

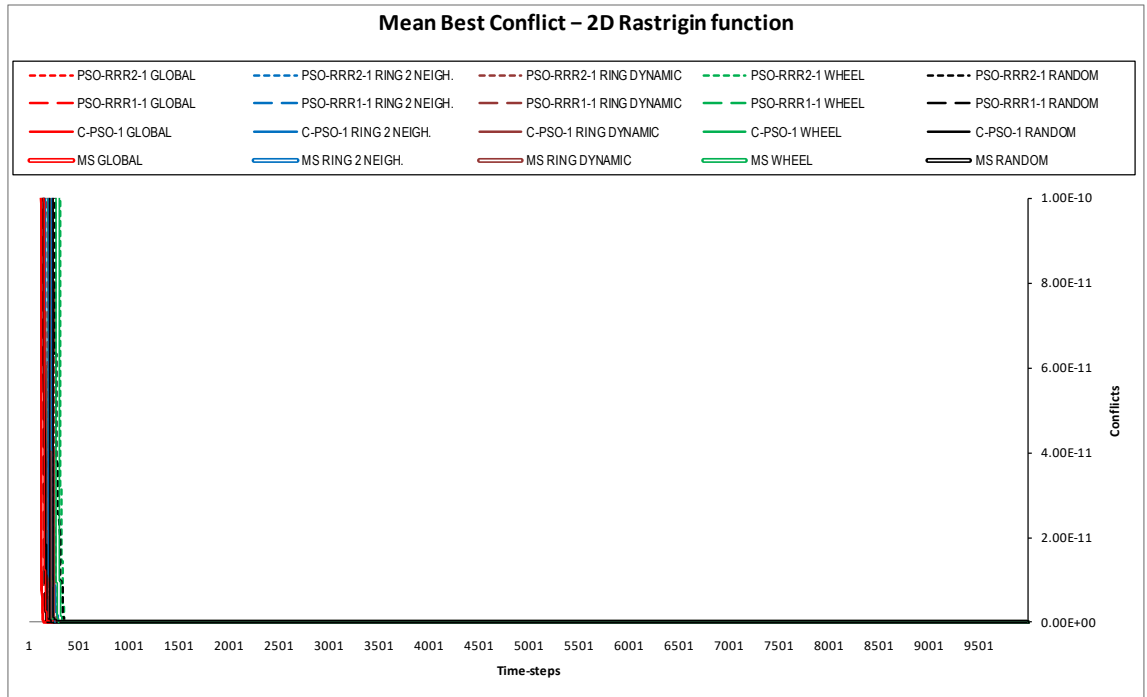

**Figure 24. Convergence curves of the mean best conflict for the 2D Schaffer f6 function, associated to Table 13. The colour-codes used to identify the neighbourhood structures are the same in the table and figure associated.**

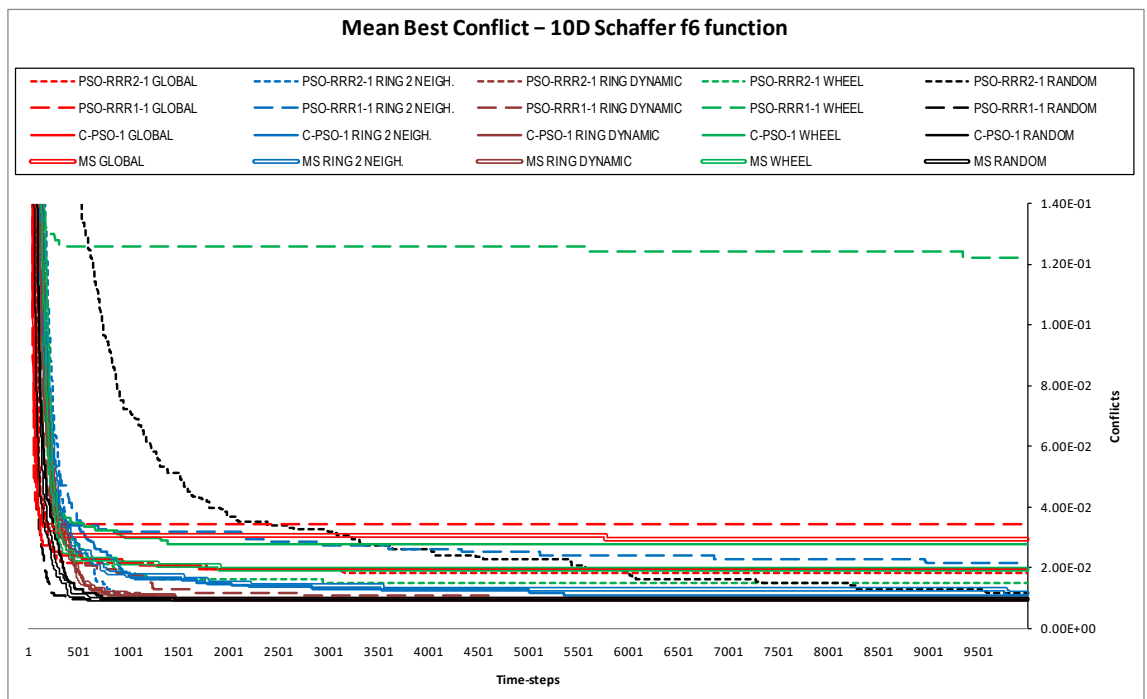

**Figure 25. Convergence curves of the mean best conflict for the 10D Schaffer f6 function, associated to Table 14. The colour-codes used to identify the neighbourhood structures are the same in the table and figure associated.**

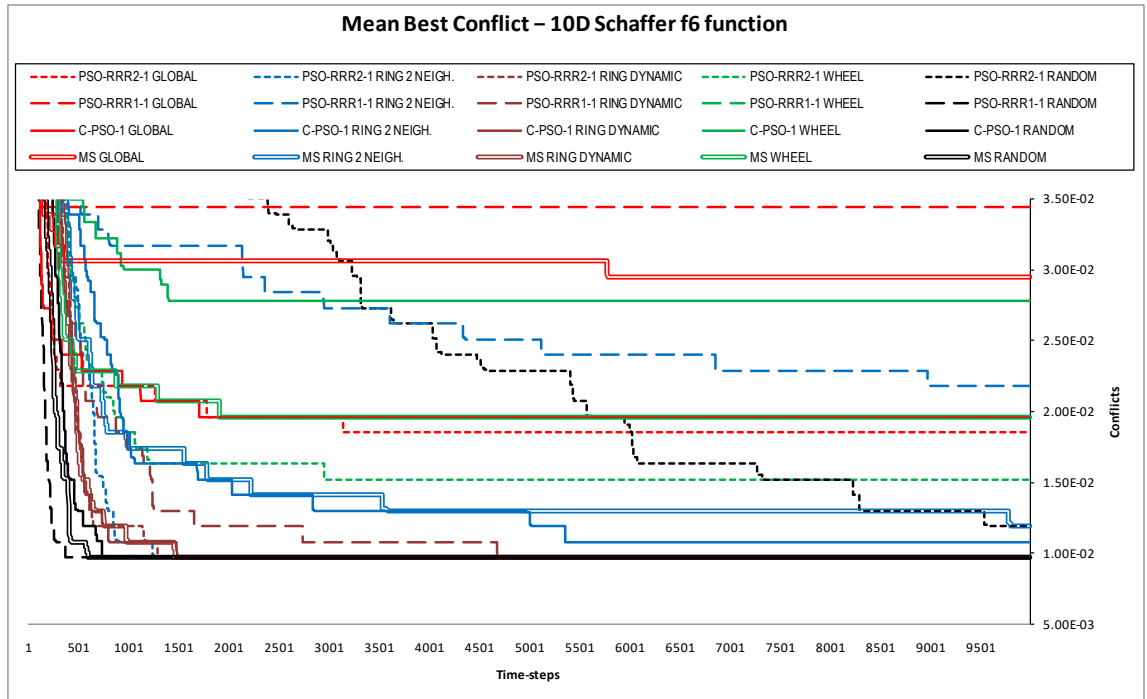

**Figure 26. Convergence curves of the mean best conflict for the 10D Schaffer f6 function, associated to Table 14. The colour-codes used to identify the neighbourhood structures are the same in the table and figure associated.**

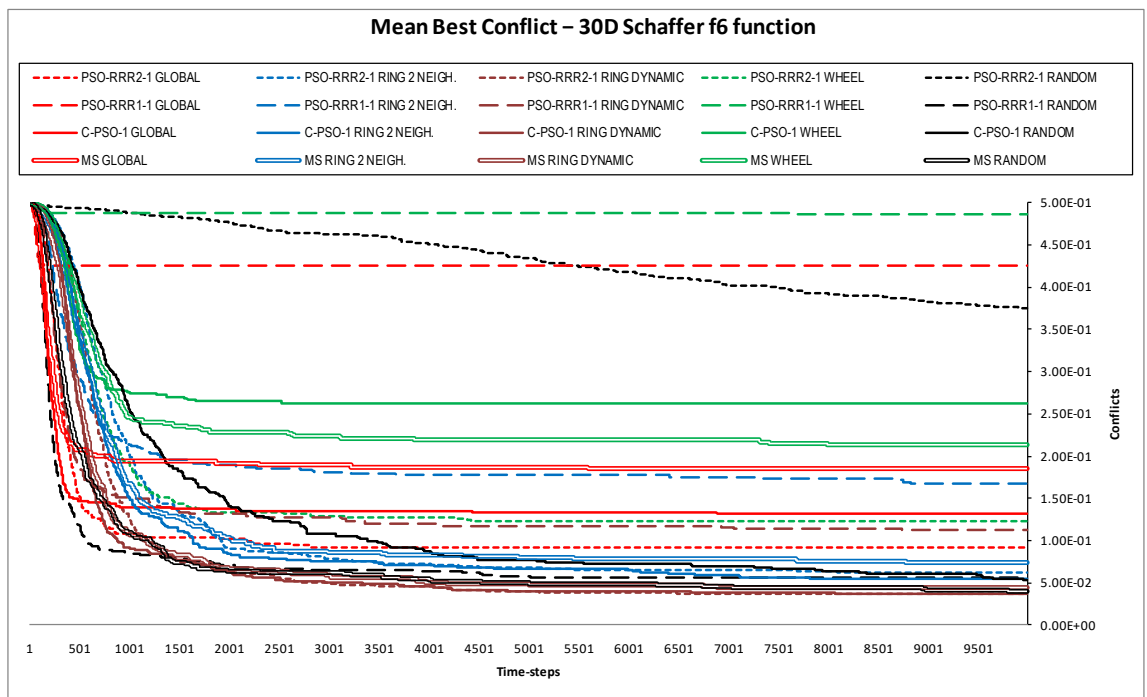

**Figure 27. Convergence curves of the mean best conflict for the 30D Schaffer f6 function, associated to Table 15. The colour-codes used to identify the neighbourhood structures are the same in the table and figure associated.**

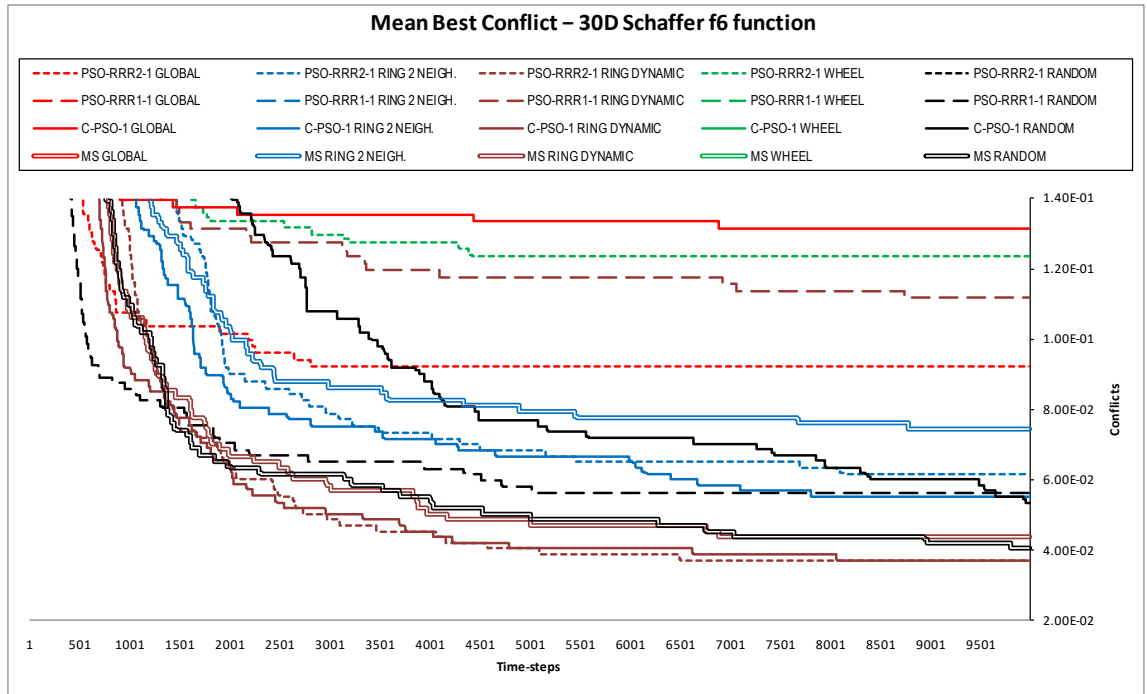

**Figure 28. Convergence curves of the mean best conflict for the 30D Schaffer f6 function, associated to Table 15. The colour-codes used to identify the neighbourhood structures are the same in the table and figure associated.**

## IV. Conclusion

The 'Ring Dynamic' topology proposed is successful and appears desirable, as the robustness gained by the reduced number of neighbours at the early stages of the search does not seem to affect the fine-grain search at the end. The performance exhibited is most of the time either between that of its global and ring ($nn$=2) counterparts –and closer to the better one– or better than both. Only a few times it happens to be worse than both, and only marginally. The global, wheel and random topologies are able to find very good results in some isolated problems, but the overall performance is inferior and remarkably less robust. The ring topology with 2 neighbours is more stable, but its performance is also less robust than that of the proposed dynamic ring topology. Therefore, for a non-problem specific optimizer, it appears desirable to implement a dynamic neighbourhood that gradually changes from very few interconnections to a highly interconnected structure as the search progresses, such as the 'Ring Dynamic' topology tested in this paper.

As to the coefficients settings working in combination with the neighbourhood structure, a similar conclusion can be drawn: the multi-swarm strategy combining different coefficients settings with different strengths and weaknesses seems to be a desirable alternative for a non-problem-specific algorithm. Settings favouring exploration such as the PSO-RRR2-1 perform really well in some cases, at the expense of bad performances in some others. The same is true for settings favouring fast convergence or even moderate speed of convergence such as the PSO-RRR1-1 and the C-PSO-1. Beware that the PSO-RRR1-1 displays a moderate to fast speed of convergence despite the high inertia weight because of the reduced randomness range (refer to Ref. 12 for a detailed discussion on the subject).